\documentclass{article} 
\usepackage{iclr2027_conference,times}

\usepackage{amsmath,amsfonts,bm}

\def\eqref#1{equation~\ref{#1}}

\def\1{\bm{1}}

\DeclareMathAlphabet{\mathsfit}{\encodingdefault}{\sfdefault}{m}{sl}
\SetMathAlphabet{\mathsfit}{bold}{\encodingdefault}{\sfdefault}{bx}{n}

\usepackage{siunitx}
\usepackage{hyperref}
\usepackage{url}
\usepackage{graphicx} 
\let\OrigIncludegraphics\includegraphics
\renewcommand{\includegraphics}[2][]{%
  \IfFileExists{#2}{\OrigIncludegraphics[#1]{#2}}{\OrigIncludegraphics[#1]{example-image}}}
\usepackage{siunitx}
\usepackage{tabularx}
\usepackage{multirow}
\usepackage{float}
\usepackage{wrapfig}
\usepackage{xcolor}
\usepackage{amsmath}
\usepackage{enumerate}
\usepackage{subcaption}
\usepackage{wrapfig}
\usepackage{colortbl}
\usepackage{diagbox}
\usepackage{makecell}
\usepackage{booktabs}

\definecolor{hdrblue}{RGB}{218,232,252}
\definecolor{rowgray}{gray}{0.93}
\title{AeroManip-VLA: Scalable Vision-Language-Action Learning for Aerial Manipulation with RL-Generated Demonstrations}

\author{Rui Huang$^{1}$, Yanlin Mu$^{2,1}$, Lidong Li$^{1}$, Yucong Wang$^{1}$, Zichen Yan$^{1}$, Lin Zhao$^{1}$ \\
$^{1}$National University of Singapore \quad $^{2}$Beijing Institute of Technology
}

\newcommand{\methodname}{AeroManip-VLA}

\iclrfinalcopy 
\begin{document}

\maketitle
\begin{figure}[h!]
    \centering
    \includegraphics[width=\linewidth]{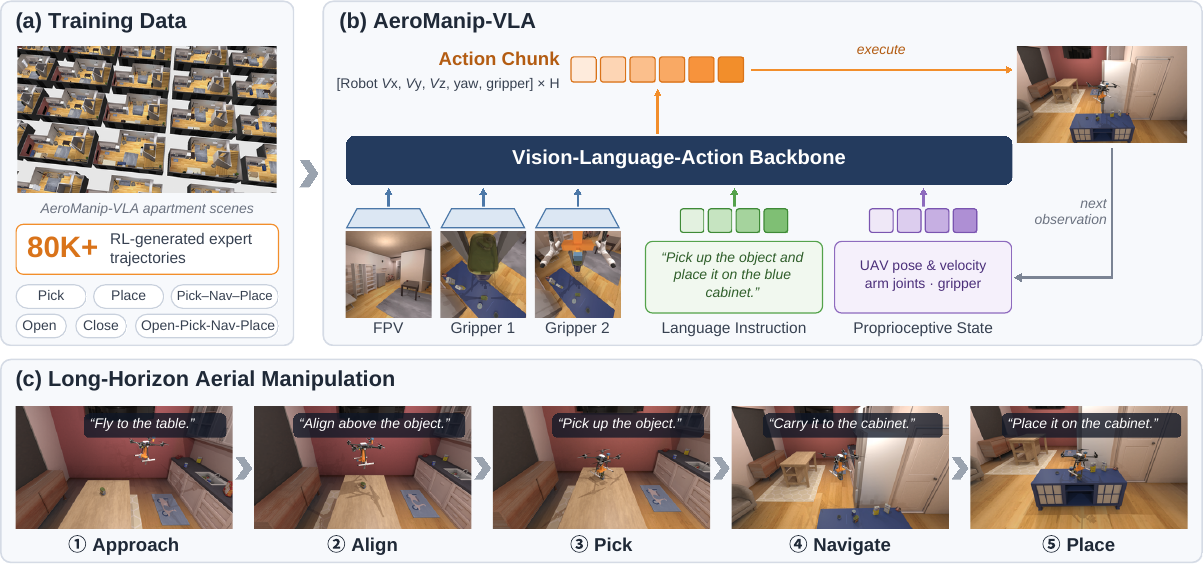}
    \caption{Overview of the AeroManip-VLA framework and dataset for aerial manipulation. Project page: \url{https://ruihuangnus.github.io/AeroManip-VLA-page/}}
    \label{fig:teaser}
\end{figure}
\begin{abstract}
Aerial manipulators extend robotic manipulation into 3D workspaces that are difficult for ground-based robots to access, creating new opportunities for general-purpose manipulation. However, extending Vision-Language-Action (VLA) models to aerial robots introduces distinct challenges due to the tight coupling between manipulation and flight, continuously changing observations, and safety-critical physical interactions. These challenges demand diverse training data and systematic policy evaluation, yet collecting demonstrations and evaluating policies directly on physical aerial platforms are costly, difficult to scale, and hard to repeat under controlled conditions.
We present \methodname, a scalable benchmark for aerial VLA data generation and policy evaluation. \methodname{} provides a GPU-accelerated simulation framework with low-level payload-aware flight and manipulation control in massively parallel environments. Building on this framework, we combine reusable reinforcement learning policies with expert task rules to automatically generate demonstrations without human teleoperation across diverse objects, environments, and randomized initial conditions. The generated data include basic skills such as grasping and placing, as well as long-horizon tasks that require both navigation and manipulation. We further introduce automated event labeling and trajectory categorization to filter demonstrations. These mechanisms enable fine-grained analysis of task progress, behavioral outcomes, and safety-related failures. Finally, we evaluate a range of imitation learning and VLA baselines across different task settings, revealing their performance characteristics and failure modes. Together, \methodname{} enables scalable aerial manipulation data generation, structured trajectory analysis, and systematic VLA evaluation in simulation prior to real-world deployment.

\end{abstract}


\section{Introduction}
Aerial manipulators offer a unique advantage over ground-based robots: their 3D mobility greatly expands the reachable workspace for general-purpose robotic manipulation, allowing them to operate at height, across obstacles, and in locations that are difficult for ground robots to access. Realizing such general-purpose capabilities requires robots to interpret diverse visual observations, understand natural-language instructions, and perform task-level reasoning. These requirements make VLA models particularly well suited for aerial manipulation. However, most existing VLA models do not target aerial manipulation, but instead focus on fixed-base or ground-based mobile manipulators, where locomotion is largely confined to planar surfaces and manipulation is performed from relatively stable platforms.

Aerial manipulation differs from ground-based manipulation in several respects. First, manipulation must be tightly coordinated with 3D flight while maintaining stability and safety. Second, the robot undergoes continuous 6-DoF motion, causing rapid changes in viewpoint and visual observations during task execution. Third, physical interactions such as contact forces, payload changes, and external disturbances can directly affect the flight dynamics ~\citep{AIR-pi}. These differences substantially broaden the operating conditions and failure modes that VLA models must handle. Accordingly, both training data and policy evaluation must capture this diversity. Scalable data generation and systematic policy evaluation are therefore essential for developing VLA systems for aerial manipulation.

Developing and evaluating VLA policies directly on physical aerial platforms, however, remains difficult. Collecting teleoperated demonstrations is costly and difficult to scale, while real-world policy evaluation is typically slow, safety-critical, and hard to repeat under controlled conditions. These practical constraints create a need for scalable simulation platforms that can generate diverse aerial manipulation data and support safe, controlled, and repeatable policy evaluation before real-world deployment.

Recent work has begun to develop scalable simulation platforms for aerial VLA~\citep{wang2025towards, mehboob2026dronevla,wang2026uav}. AIR-VLA~\citep{Air-vla} introduced the first benchmark specifically designed for VLA-based aerial manipulation, providing a physics-based simulation environment and a multimodal dataset of 3,000 manually teleoperated demonstrations. The reliance of AIR-VLA on human teleoperation limits the scalability of data collection. Extending the dataset to new objects, scene configurations, initial poses, or task variations requires additional manual demonstrations, even when the basic manipulation skills, such as grasping and placing, remain largely unchanged. Manual data collection makes it difficult to systematically cover diverse aerial manipulation scenarios and limits dataset expansion as tasks and environments change. These limitations motivate our development of a scalable framework that can automatically generate diverse aerial manipulation demonstrations, characterize their quality and failure modes, and support systematic policy evaluation across a broad range of task conditions.

We propose \methodname, a scalable benchmark for aerial VLA data generation and policy evaluation. First, we develop a GPU-accelerated simulation framework that supports low-level flight and manipulation control in massively parallel environments. Built on this framework, expert rules and reinforcement learning policies are combined to automatically generate demonstrations without human teleoperation, covering diverse objects, environments, and randomized initial conditions. The generated data include both basic manipulation skills and long-horizon tasks requiring coordinated navigation and physical interaction. Second, we introduce automated event labeling and trajectory categorization to organize demonstrations according to task progress, behavioral outcomes, and safety-related events. This enables systematic data filtering and analysis of VLA failure modes. Finally, we evaluate a range of imitation-learning and VLA baselines under controlled task settings, providing reference results for aerial manipulation research. Together, these components form a scalable pipeline for aerial manipulation data generation, policy evaluation, and simulation-based validation prior to real-world deployment.

In summary, this paper makes the following contributions:

\textbf{A scalable GPU-accelerated benchmark for aerial VLA.}
To the best of our knowledge, \methodname ~is the first aerial VLA benchmark to combine massively parallel simulation, automated demonstration generation, and trajectory-level analysis within a unified framework. It supports low-level payload-aware flight and manipulation control, enabling efficient data generation, policy training, and evaluation across diverse aerial manipulation tasks.

\textbf{Scalable demonstration generation without human teleoperation.}
We combine expert rules with reinforcement learning policies to automate demonstration generation, removing the need for manual teleoperation and enabling scalable data collection across different objects, environments, and randomized initial poses. The resulting dataset covers both basic manipulation skills and long-horizon tasks that require coordinated navigation and manipulation.

\textbf{Structured trajectory analysis and comprehensive baseline evaluation.}
We develop an automated event-labeling and trajectory-categorization scheme for fine-grained analysis of task progress, behavioral outcomes, and safety-related failures. This analysis goes beyond aggregate success rates. We further evaluate a range of imitation learning (IL) and VLA baselines across different task settings, revealing their performance characteristics and failure modes while providing reference results for future aerial manipulation research.

\vspace{-3mm}
\section{Related Work}
\vspace{-2mm}

\textbf{Benchmarks and Simulation for Robot Learning:}
LIBERO~\citep{LIBERO} and BEHAVIOR-1K~\citep{li2023behavior} provide diverse manipulation and long-horizon task suites. ManiSkill3~\citep{taomaniskill3} enables GPU-parallel simulation, and ManiSkill-HAB~\citep{maniskill-hab} further integrates low-level RL/IL baselines and trajectory filtering for controlled data generation. RoboCasa365~\citep{nasiriany2026robocasa365} expands household task and environment diversity, whereas RoboVerse~\citep{geng2025roboverse} and RoboTwin 2.0~\citep{chen2025robotwin} integrate synthetic data generation, domain randomization, and standardized evaluation. These benchmarks primarily target ground-based robots, leaving coupled flight--manipulation dynamics and aerial safety constraints underexplored.
Recent benchmarks have begun to consider aerial manipulation. AIR-VLA~\citep{Air-vla} provides a physics-based simulator, 3,000 teleoperated demonstrations, and VLA/VLM evaluation. AM-Bench~\citep{AM-Bench} further introduces a modular aerial manipulation benchmark with diverse robot embodiments, low-level controllers, physical disturbances, and policy-learning baselines across 12 manipulation tasks. However, scalable generation and analysis of diverse aerial VLA demonstrations remain underexplored, especially for long-horizon tasks involving randomized objects, environments, and initial conditions.

\textbf{VLA Models for Aerial Manipulation:}
Recent studies have begun to extend VLA models to aerial robots. DroneVLA~\citep{mehboob2026dronevla} combines language-conditioned perception with aerial navigation and handover. $\pi$, But Make It Fly~\citep{AIR-pi} adapts manipulation-pretrained VLAs to real aerial robots using synthetic navigation data, and AIR-VLA+~\citep{AIR-VLA+} improves the integration of flight and manipulation within the VLA architecture. Despite these advances, scalable generation of diverse \emph{manipulation} demonstrations remains comparatively underexplored. Existing aerial VLA datasets still rely heavily on manually collected trajectories, whereas synthetic data generation has primarily focused on navigation. Our work targets this remaining gap through automated aerial manipulation data generation, together with systematic trajectory analysis and policy evaluation.

\textbf{Scalable Demonstration Generation and Skill Composition:}
The high cost of human teleoperation has motivated automated demonstration generation. MimicGen~\citep{mandlekar2023mimicgen} adapts human demonstrations to new scene configurations, DexMimicGen~\citep{jiang2025dexmimicgen} extends this idea to bimanual dexterous manipulation, and DemoGen~\citep{xue2025demogen} augments trajectories and visual observations through spatial transformations. For long-horizon tasks, SkillMimicGen~\citep{garrett2024skillmimicgen} composes reusable skills through planned transit motions, and LodeStar~\citep{raghunandan2022lodestar} combines demonstration augmentation, reinforcement learning, and skill composition.
Despite these advances, most existing methods still depend on human source demonstrations and primarily target fixed-base or ground-based manipulation. Aerial manipulation additionally requires coordinating 3D flight with physical interaction across varying initial states and task configurations. Our \methodname instead combines reusable closed-loop reinforcement learning policies with expert task rules to automatically generate demonstrations across randomized objects, environments, and initial conditions without human teleoperation. These skills are further composed into long-horizon demonstrations that integrate navigation and manipulation for aerial VLA training and evaluation.

\section{Methodology}
\subsection{Benchmark Design}
The benchmark comprises four task settings.
\textbf{TidyHouse} and \textbf{PrepareGroceries} provide
object-specific Pick and Place tasks for household and kitchen
rearrangement, respectively.
\textbf{PackageDelivery} extends the benchmark to outdoor delivery, requiring the aerial manipulator to pick up a package from a vehicle, navigate to the entrance of a designated room, and place the package on top of a cabinet by the door.
\textbf{LongHorizonRearrangement} evaluates continuous single-object rearrangement tasks combining long-range navigation, pick-and-place,
and opening and closing kitchen cabinet and refrigerator doors.

\subsubsection{Task.}
\textbf{Basic Manipulation skill Definitions.}
Our benchmark defines five parameterized skills for aerial manipulation: \emph{Pick}, \emph{Place}, \emph{Open}, \emph{Close}, and \emph{Nav}.
Given an object $x$ at pose $x_{\mathrm{pose}}$, $\mathrm{\textbf{Pick}}[a](x_{\mathrm{pose}})$ picks up the object, while $\mathrm{\textbf{Place}}[a](x_{\mathrm{pose}}, g_{\mathrm{pos}})$ places it at the goal position $g_{\mathrm{pos}}$.
For these two skills, the optional parameter $a$ specifies the articulated structure from which the object is picked or into which it is placed.
The skills $\mathrm{\textbf{Open}}[a](a_{\mathrm{pos}})$ and $\mathrm{\textbf{Close}}[a](a_{\mathrm{pos}})$ respectively open and close articulation $a$ by interacting with its handle at position $a_{\mathrm{pos}}$.
Finally, $\mathrm{\textbf{Nav}}(p_{\mathrm{target}})$ moves the aerial manipulator to a specified target position $p_{\mathrm{target}}$.
These skills serve as building blocks for long-horizon tasks that combine navigation and manipulation.

\textbf{Long-Horizon Manipulation.}
We compose the basic skills into continuous single-object trajectories.
The core \textsc{PickPlace} sequence is defined as
\begin{equation}
    \mathrm{PickPlace}(x,g)
    =
    \mathrm{Pick}(x_{\mathrm{pose}})
    \rightarrow
    \mathrm{Nav}(p_{\mathrm{target}})
    \rightarrow
    \mathrm{Place}(\langle x_{\mathrm{pose}}, g_{\mathrm{pos}} \rangle),
\end{equation}
where navigation transports the grasped object toward the placement goal.
\textbf{PackageDelivery} follows this sequence, while
\textbf{LongHorizonRearrangement} additionally incorporates
\emph{Open} and \emph{Close} at task-dependent stages to access or
store objects in cabinets and refrigerators.
All skills are executed continuously within a single episode,
requiring precise contact-rich manipulation while maintaining
flight stability.

\subsubsection{Simulator.} 
We build \methodname{} on ManiSkill3~\citep{taomaniskill3}, using
ReplicaCAD indoor environments and task settings adapted from
 ManiSkill-HAB~\citep{maniskillhab}
and HAB~\citep{habitat2.0}. The simulator supports GPU-parallel physics and rendering, enabling
large numbers of aerial manipulation environments to be executed
concurrently for scalable data generation, policy training, and
evaluation.

\subsubsection{Robot System.}
Unlike AIR-VLA~\citep{Air-vla}, which uses a 7-DoF Franka Panda
manipulator weighing approximately 18\,kg, we focus on lightweight
manipulators that better reflect the payload constraints of small
quadrotors. Following Air-UMI~\citep{Air-UMI}, our simulated platform
is based on an X500 quadrotor and supports two configurations: a
lightweight 1-DoF gripper platform and a 3-DoF arm, both equipped with
UMI-style fingertips~\citep{UMI}. Together, they provide complementary
levels of reach and dexterity for aerial manipulation.
We use a payload-aware cascaded flight controller to track world-frame velocity and yaw-rate commands. The controller also supports batched GPU execution across parallel simulation environments.
Grasping and object transport are realized through simulated physical contact; no rigid object attachment or pose teleportation is used during task execution.

\subsubsection{Observation and Action Spaces.}
The policy receives three onboard RGB-D views
(\texttt{drone front}, \texttt{drone hand}, and \texttt{drone own})
together with a 19-dimensional proprioceptive state. Language-conditioned policies additionally receive a natural-language
instruction. 
The policy outputs a normalized five-dimensional continuous action
\[
a_t=[u_x,u_y,u_z,u_\psi,u_g]\in[-1,1]^5,
\]
representing commands for world-frame translational velocity, yaw rate, and gripper opening.

\subsection{Payload-aware Control for Aerial Manipulation}
\label{sec:method:control}
We organize aerial manipulation into a shared low-level control
layer and a task-level policy layer.
The low-level controller executes flight commands in GPU-parallel
environments and compensates for payload changes during grasping,
transport, and release.
On top of this controller, we construct a privileged hybrid expert
for demonstration generation and train skill policies using
behavior cloning and reinforcement learning.
All task-level policies share the same low-level flight controller.

\textbf{GPU-parallel flight.}
To simulate many flying robots in parallel on the GPU, we represent each free-flying airframe using six virtual joints attached to the world frame. Three prismatic joints describe translation along the world $x$-, $y$-, and $z$-axes, while three rotational joints describe yaw, pitch, and roll. The rotor force and torque, denoted by $W=(F,\tau_O)$ and expressed in the world frame, are converted into forces and torques acting on these virtual joints through $Q=J^{\top}W$, where $J$ is the corresponding airframe Jacobian. Physics and low-level control are updated at $240\,\mathrm{Hz}$, while the policy runs at $20\,\mathrm{Hz}$. The airframe state is refreshed at every physics step. In a $25\,\mathrm{s}$ flight test, the GPU simulation differs from the CPU floating-base reference by only $2$--$3\,\mathrm{mm}$.

\textbf{Cascaded flight control.}
We use a cascaded velocity--attitude controller to track the world-frame
translational-velocity command $v^{\mathrm{cmd}}$ and yaw-rate command
$\dot{\psi}^{\mathrm{cmd}}$ generated by the task-level policy.
Physics and low-level control are updated at
$\Delta t=1/240\,\mathrm{s}$.
The outer-loop velocity controller computes
\begin{equation}
    e_v
    =
    v^{\mathrm{cmd}}-v,
    \qquad
    \xi
    \leftarrow
    \operatorname{clip}
    \left(
        \xi+e_v\Delta t,
        -2,
        2
    \right),
    \label{eq:vel}
\end{equation}
where $e_v$ is the velocity tracking error and $\xi$ is its integral term.
Here, $\operatorname{clip}(x,l,u)$ denotes componentwise saturation of
$x$ to the interval $[l,u]$, preventing excessive accumulation of the
integral term.
The desired world-frame force is then
\begin{equation}
    f^{\mathrm{des}}
    =
    \hat m
    \left(
        \operatorname{clip}_{\mathrm{acc}}
        \left(
            3e_v+\xi
        \right)
        +
        g\mathbf e_z
    \right),
    \label{eq:force}
\end{equation}
where $\hat m$ is the estimated total mass, including the payload when
present, and $\mathbf e_z$ is the upward unit vector.
The operator $\operatorname{clip}_{\mathrm{acc}}$ limits the commanded
acceleration to $\pm 3\,\mathrm{m/s^2}$ along each horizontal axis and
$\pm 5\,\mathrm{m/s^2}$ vertically.
The commanded yaw rate is integrated to obtain the desired heading,
\begin{equation}
    \psi^{\mathrm{des}}
    \leftarrow
    \psi^{\mathrm{des}}
    +
    \dot{\psi}^{\mathrm{cmd}}\Delta t.
\end{equation}
The desired attitude $R^{\mathrm{des}}$ is constructed from
$f^{\mathrm{des}}$ and $\psi^{\mathrm{des}}$, with its body $z$-axis
aligned with the desired force direction.
The corresponding desired angular velocity is denoted by
$\omega^{\mathrm{des}}$.
The inner-loop attitude controller computes the body torque as
\begin{equation}
    \tau
    =
    -k_R e_R(R,R^{\mathrm{des}})
    -k_\omega(\omega-\omega^{\mathrm{des}}),
    \label{eq:att}
\end{equation}
where $e_R$ is the $SO(3)$ attitude error.
The total thrust is obtained by projecting the desired force onto the
current body $z$-axis,
\begin{equation}
    T
    =
    (f^{\mathrm{des}})^\top
    R\mathbf e_z.
\end{equation}
Finally, the total thrust and body torque are allocated to the four rotors as
\begin{equation}
    f_{\mathrm{rotor}}
    =
    \operatorname{clip}
    \left(
        A^{-1}
        \begin{bmatrix}
            T\\
            \tau
        \end{bmatrix},
        0,
        f_{\max}
    \right),
\end{equation}
where $A$ is the control-allocation matrix and $f_{\max}$ is the
per-rotor thrust limit.
The resulting rotor wrench is transformed to the world frame and applied
through the virtual joints.
For the unloaded platform, $\hat m=m_0$, and the attitude-control gains are
set to $(k_R,k_\omega)=(2.5,0.55)$.

\textbf{Payload-aware compensation.}
After grasp detection, the controller updates the feedforward
mass using the simulator-provided object mass and increases
attitude gains.
Once the placement support carries the object, it restores
the unloaded mass and resets the vertical integral to avoid
excess thrust after release.
Compensation is enabled for selected object classes, and all
evaluated policies share the same low-level controller.

\subsection{Policy Learning for Aerial Manipulation}
We consider three complementary approaches for generating aerial
manipulation behavior. Unlike AIR-VLA~\citep{Air-vla}, which relies on
manually teleoperated demonstrations, our pipeline emphasizes automated
and scalable data generation. The privileged hybrid expert provides
reliable demonstrations using simulator knowledge and structured
control. Behavior cloning distills these demonstrations into standalone
learned policies, while PPO further optimizes the policies through
environment interaction and task rewards. Together, these approaches
enable reproducible and systematic data generation across diverse task
configurations.

\subsubsection{Privileged Hybrid Expert}
\label{sec:method:expert}

We construct a privileged hybrid teacher for automated demonstration
generation. The teacher combines frozen learned components, rule-based
control, geometric planning, and scripted feedback, with access to
privileged simulator information such as object poses, collision
geometry, and task-specific targets.

Geometric planning determines feasible initial configurations and
collision-free motion when global spatial reasoning is required.
During local interaction, learned components provide mode selection
and feedback control, while rule-based guards enforce geometric,
contact, and stability constraints. Scripted controllers are used for
stages that can be reliably specified from privileged task states.

The teacher is used only to generate successful demonstrations and
initialization resources. It is not itself an evaluated policy and
does not involve reinforcement-learning updates.

\subsubsection{Skill Policy Learning}
\label{sec:method:learning}

Given demonstrations generated by the privileged teacher, we study
learned skill policies using behavior cloning (BC) and reinforcement
learning.

\textbf{Behavior Cloning.}
We train BC policies to imitate the continuous expert actions from
successful demonstrations by minimizing the mean squared error between
predicted and normalized actions. Unlike the frozen BC components
inside the privileged teacher, these policies directly predict the
complete continuous control action and serve as standalone learned
skill policies.

\textbf{PPO Training.}
We further optimize skill policies with
PPO in GPU-parallel environments.
The policy uses a Gaussian actor with a separate critic, generalized
advantage estimation, and clipped policy updates. Reward functions are
defined according to task progress and success criteria, together with
penalties for undesirable actions and contacts.
PPO policies can be trained either from scratch or initialized from a
pretrained BC policy. When BC initialization is used, optional critic
warm-up and imitation regularization can be applied during early
training. The reference configuration uses 64-step rollout windows,
$\gamma=0.97$, and a clipping parameter of $0.2$.
The learned skill policies are then composed with privileged geometric
planning and low-level scripted control to generate task-level rollouts.
This pipeline is applied consistently across all manipulation skills.

\subsection{Event Annotation for Aerial Manipulation}
\label{sec:method:annotation}

We develop an event-based annotation protocol tailored to the coupled
requirements of aerial flight and manipulation. Annotations combine
task-specific interaction events with aerial operating constraints,
including excessive tilt, altitude violations, non-target contact,
payload loss, platform stability, and clearance conditions. This
enables task completion to be assessed jointly with the physical
constraints specific to aerial manipulation.

We retain both skill-level outcomes and full-trajectory outcomes.
A successfully completed skill remains valid even if a subsequent stage
fails, while time-limit truncations are distinguished from physical
failures. Parent-trajectory identifiers and event boundaries preserve
the relationship between extracted skill segments and their original
rollouts.

We additionally verify the validity of skill transitions when
constructing segmented demonstrations. In particular, downstream
segments must begin from eligible states that satisfy the required
handoff conditions rather than from states already close to task
completion. These annotations provide traceable criteria for
trajectory segmentation, demonstration selection, and structured
failure analysis.

\section{Experiments}
\label{sec:experiments}

We organize our evaluation around two questions.
\textbf{First}, how do expert-, imitation-, and reinforcement-learning-based
data generation strategies compare in success rate, learning efficiency,
trajectory diversity, and execution efficiency?
\textbf{Second}, how well do representative visuomotor and vision-language-action
policies learn aerial manipulation from the generated demonstrations?
\label{sec:experiments:setup}
\begin{figure}[t]
    \centering
    \includegraphics[width=\linewidth]{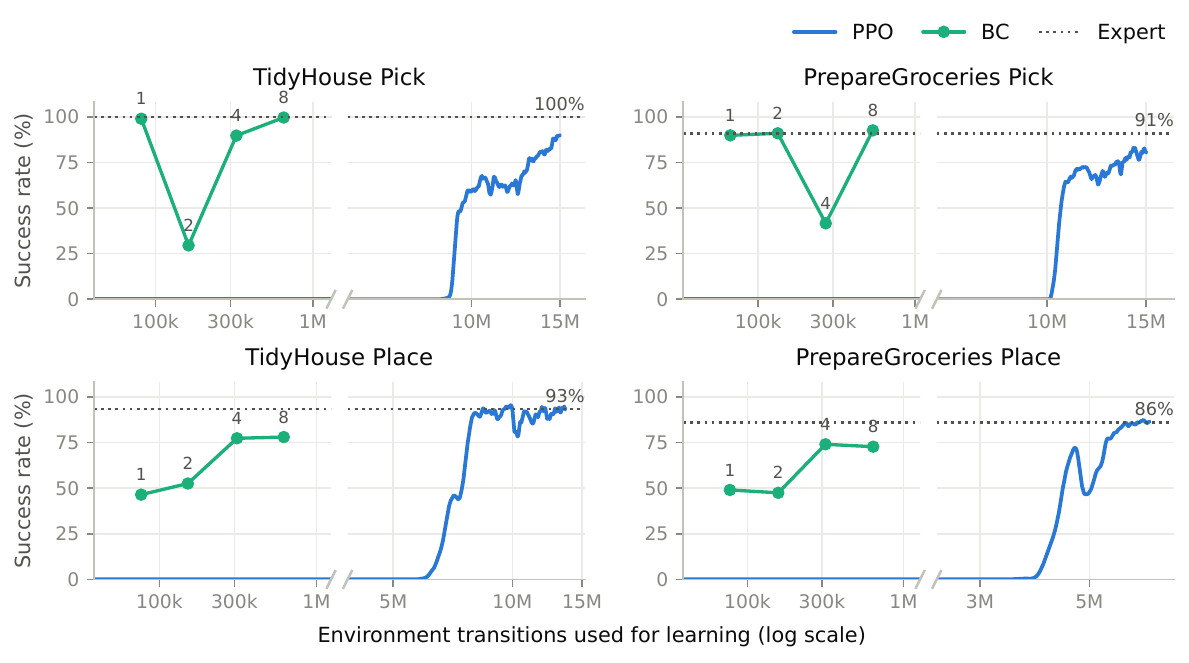}
    \vspace{-7mm}
    \caption{\textbf{BC is substantially more interaction-efficient than PPO, but relies on expert demonstrations.}
BC markers report held-out success after training on successful expert
trajectories from 1, 2, 4, and 8 collection rounds (256 rollouts each).
PPO curves show training success from scratch (400-episode moving average),
while dotted lines denote expert held-out success.
The broken log-scale $x$-axis reports demonstration transitions for BC
and environment interactions for PPO. One training seed per method.}
    \label{fig:learning_curves}
\vspace{-5mm}
\end{figure}

\subsection{Evaluation of Data Generation Strategies}
\label{sec:experiments:data_generation}
\textbf{Success rates and learning efficiency.}
Figure~\ref{fig:learning_curves} shows that learning efficiency alone
does not favor PPO as a replacement for expert-guided collection.
On Pick, BC approaches expert success with only 67--81k demonstration
transitions, whereas PPO requires millions of interactions and
remains below the expert in held-out evaluation.
These results support the hybrid expert as the source of our
released dataset and BC as a low-cost route to policy distillation.
Our motivation for evaluating PPO is therefore different:
whether reward-driven learning can produce successful trajectories
with greater diversity or lower collection cost.
The following analyses examine whether these benefits can justify
its higher upfront training cost.

\textbf{Trajectory diversity.}
\begin{figure}[t]
  \centering
  \includegraphics[width=\linewidth]{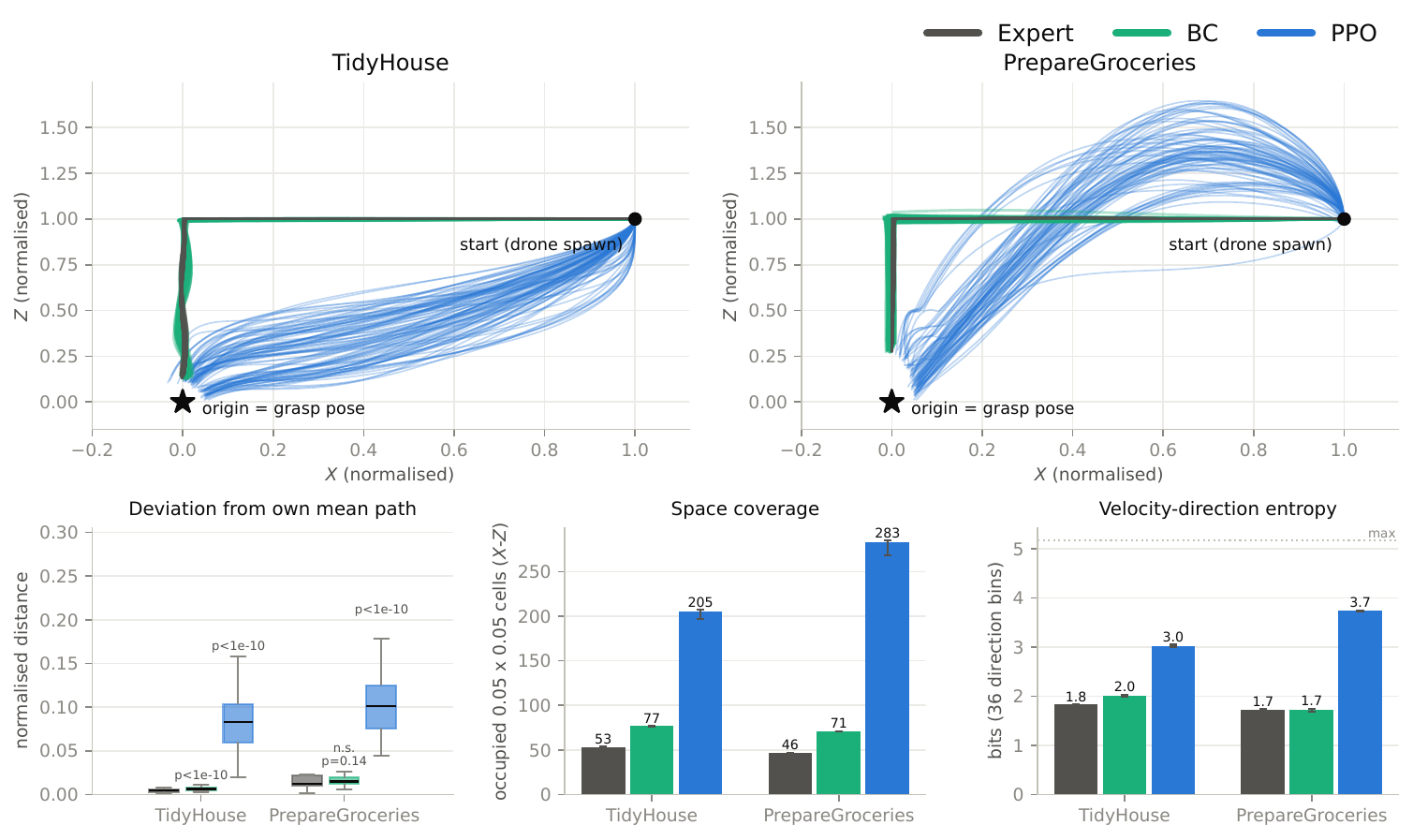}
  \vspace{-7mm}
  \caption{\textbf{PPO produces substantially more diverse successful approach trajectories than the expert and BC.}
    Top: successful approach segments from drone spawn to within 5\,cm
    of the planned grasp pose, shown in a start-normalised frame with
    the grasp pose at the origin and all trajectories starting at $(1,1)$.
    Expert and BC follow a similar L-shaped approach, whereas PPO
    spreads across the $X$--$Z$ plane.
    Bottom: deviation from each policy's own mean path, spatial coverage
    measured by occupied $0.05 \times 0.05$ cells, and velocity-direction
    entropy over 36 bins.
    Coverage and entropy are estimated from 200 random subsets of
    80 episodes per policy; bars show the mean and 95\% interval.}
    \label{fig:trajectory_diversity}
    \vspace{-4mm}
\end{figure}
While Figure~\ref{fig:learning_curves} shows that PPO is much less
interaction-efficient than BC, Figure~\ref{fig:trajectory_diversity}
reveals what this additional optimization can buy in return.
Across both tasks, PPO reaches the grasp pose through a much broader
set of successful approach trajectories.
The expert and BC largely follow the same near-deterministic
L-shaped strategy (navigate $\rightarrow$ descend onto the object), whereas PPO explores many alternative routes in the
normalised $X$--$Z$ plane.

This difference is reflected in all three diversity metrics.
PPO exhibits substantially larger deviation from its own mean path,
indicating greater variation across successful rollouts.
It also covers far more spatial cells
(TidyHouse: 205 vs.\ 53 for Expert and 77 for BC;
PrepareGroceries: 283 vs.\ 46 and 71),
and achieves markedly higher velocity-direction entropy
(TidyHouse: 3.0 bits vs.\ 1.8 and 2.0;
PrepareGroceries: 3.7 bits vs.\ 1.7 and 1.7).
These results suggest that, although PPO is not a sample-efficient
replacement for expert-guided collection, it can uncover successful
approach behaviors that are largely absent from expert and BC data.

\begin{figure}[t]
  \centering
  \includegraphics[width=0.9\linewidth]{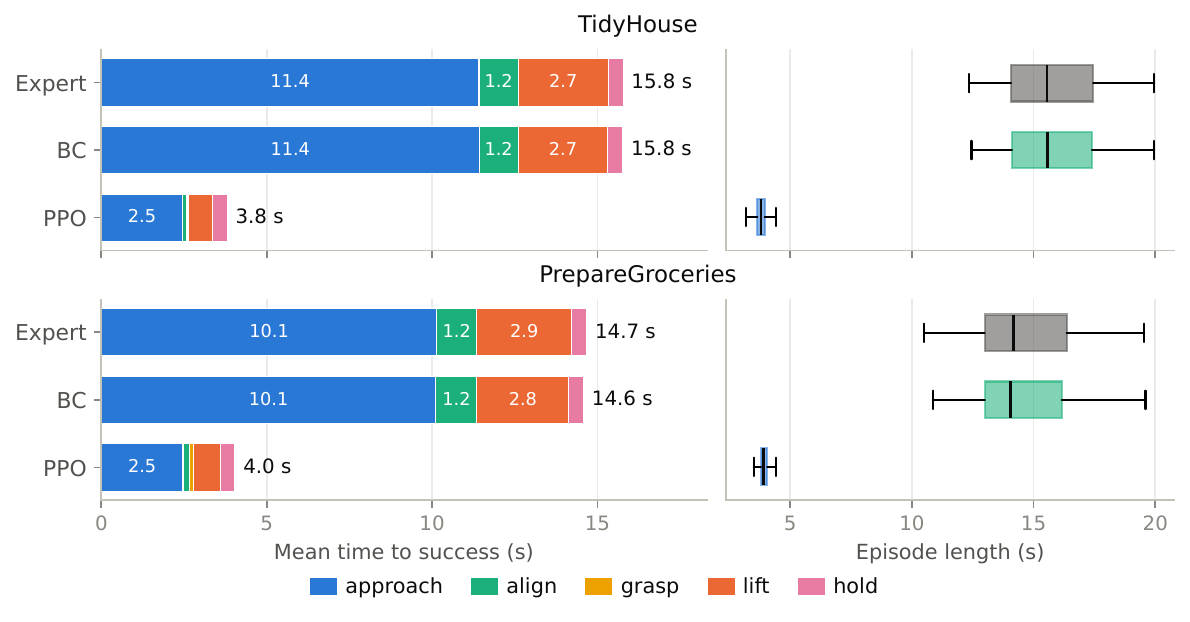}
  \vspace{-5mm}
  \caption{\textbf{PPO discovers diverse yet substantially faster successful trajectories.}
Left: mean time to success decomposed into approach, alignment,
grasp, lift, and hold phases.
PPO reduces total execution time from 15.8\,s to 3.8\,s on
TidyHouse and from approximately 14.7\,s to 4.0\,s on
PrepareGroceries, primarily through a shorter approach phase.
Right: distribution of successful episode lengths for each policy.}
  \label{fig:ppo_efficiency}
  \vspace{-2mm}
\end{figure}

\textbf{Execution efficiency.}
The increased diversity of PPO does not come from simply taking longer
or more circuitous routes.
As shown in Figure~\ref{fig:ppo_efficiency}, successful PPO rollouts are
substantially faster than those of the expert and BC.
On TidyHouse, PPO reduces the mean time to success from 15.8\,s
for both the expert and BC to 3.8\,s; on PrepareGroceries, it reduces
14.7--14.6\,s to 4.0\,s.
Most of this improvement comes from the approach phase, which decreases
from 11.4\,s to 2.5\,s on TidyHouse and from 10.1\,s to 2.5\,s on
PrepareGroceries.
Together with Figure~\ref{fig:trajectory_diversity}, these results show
that PPO discovers alternative approach strategies that are not only
more diverse, but also substantially more efficient to execute. This gain should be distinguished from learning efficiency:
PPO requires much more interaction to train, but once trained, its
successful trajectories are considerably shorter.
Since these statistics are conditioned on successful episodes, they
measure execution efficiency rather than end-to-end demonstration
collection throughput.

Overall, BC efficiently reproduces expert behavior, whereas PPO requires
substantially more training interaction but discovers a broader set of
successful strategies. Importantly, these successful trajectories are
not merely more diverse; they are also substantially faster to execute.

\subsection{Downstream Policy Benchmarking}
\label{sec:experiments:benchmark}

\textbf{Baseline methods.}
We evaluate four representative policies on our aerial manipulation dataset, covering both visuomotor imitation learning and pretrained vision-language-action (VLA) models. ACT~\cite{ACT} uses a Transformer-based conditional variational autoencoder to predict action chunks, while Diffusion Policy
(DP)~\cite{DP} generates action sequences through conditional denoising to model multimodal action distributions. For pretrained VLA baselines, we include $\pi_0$~\cite{pi0}
and $\pi_{0.5}$~\cite{pi0.5}, which generate continuous actions using flow matching.
Together, these baselines allow us to assess how effectively different policy architectures learn from our generated
demonstrations and how pretrained VLAs transfer to aerial manipulation.

\textbf{Overall performance.}
The benchmark clearly differentiates the evaluated policy classes.
Among the four baselines, $\pi_{0.5}$ achieves the highest mean success
rate of $44.6\%$, followed by $\pi_0$ at $38.9\%$, DP at $33.9\%$,
and ACT at $29.5\%$. The stronger performance of the pretrained VLA
models indicates effective transfer to aerial manipulation, while the
remaining gap to full task completion leaves substantial room for
future methods.

\begin{table*}[t]
\centering
\small

\caption{
\textbf{Simulation results on \methodname{} PrepareGroceries.}
For each skill--object pair, the table reports the success rate (\%)
over evaluation episodes together with its binomial standard error.
Mean S.R. is the unweighted average over all 16 Pick/Place--object
combinations.
}
\vspace{-2mm}
\label{tab:pg_pick_place_hab}

\begingroup
\setlength{\tabcolsep}{3pt}
\setlength{\arrayrulewidth}{0.3pt}
\renewcommand{\arraystretch}{1.15}
\arrayrulecolor{black!50}

\resizebox{\textwidth}{!}{%
\begin{tabular}{ccccccccccc}
\Xhline{0.8pt}

\diagbox[width=1.5cm,height=0.95cm]{Alg.}{Obj.}
&
\emph{Skill}
&
\makecell{Master Chef\\Can}
&
\makecell{Sugar\\Box}
&
\makecell{Tomato\\Soup Can}
&
\makecell{Tuna Fish\\Can}
&
\makecell{Pudding\\Box}
&
\makecell{Gelatin\\Box}
&
\makecell{Potted\\Meat Can}
&
Bowl
&
\makecell{Mean S.R.\\(Pick + Place)}
\\
\Xhline{0.5pt}

& Pick
& $2.0 \pm 2.0$
& $0.0 \pm 0.0$
& $0.0 \pm 0.0$
& $0.0 \pm 0.0$
& $0.0 \pm 0.0$
& $0.0 \pm 0.0$
& $10.0 \pm 4.2$
& {\boldmath$18.0 \pm 5.4$}
& \\

\multirow{-2}{*}{ACT}
& Place
& $24.0 \pm 6.0$
& {\boldmath$98.0 \pm 2.0$}
& $28.0 \pm 6.3$
& $82.0 \pm 5.4$
& $46.0 \pm 7.0$
& $48.0 \pm 7.1$
& $56.0 \pm 7.0$
& {\boldmath$60.0 \pm 6.9$}
& \multirow{-2}{*}{29.5}
\\
\hline

& Pick
& $4.0 \pm 2.8$
& $2.0 \pm 2.0$
& $6.0 \pm 3.4$
& $4.0 \pm 2.8$
& $2.0 \pm 2.0$
& $2.0 \pm 2.0$
& $2.0 \pm 2.0$
& $6.0 \pm 3.4$
& \\

\multirow{-2}{*}{DP}
& Place
& $50.0 \pm 7.1$
& $90.0 \pm 4.2$
& $50.0 \pm 7.1$
& $78.0 \pm 5.9$
& $82.0 \pm 5.4$
& $80.0 \pm 5.7$
& $32.0 \pm 6.6$
& $52.0 \pm 7.1$
& \multirow{-2}{*}{33.9}
\\
\hline

& Pick
& $6.7 \pm 4.6$
& {\boldmath$6.7 \pm 4.6$}
& {\boldmath$10.0 \pm 5.5$}
& {\boldmath$20.0 \pm 7.3$}
& $13.3 \pm 6.2$
& $13.3 \pm 6.2$
& $13.3 \pm 6.2$
& $6.7 \pm 4.6$
& \\

\multirow{-2}{*}{PI0}
& Place
& {\boldmath$80.0 \pm 5.7$}
& $82.0 \pm 5.4$
& $52.0 \pm 7.1$
& $88.0 \pm 4.6$
& $52.0 \pm 7.1$
& $68.0 \pm 6.6$
& {\boldmath$86.0 \pm 4.9$}
& $24.0 \pm 6.0$
& \multirow{-2}{*}{38.9}
\\
\hline

& Pick
& {\boldmath$8.0 \pm 3.8$}
& $6.0 \pm 3.4$
& $6.0 \pm 3.4$
& $12.0 \pm 4.6$
& {\boldmath$14.0 \pm 4.9$}
& {\boldmath$16.0 \pm 5.2$}
& {\boldmath$16.0 \pm 5.2$}
& $14.0 \pm 4.9$
& \\

\multirow{-2}{*}{PI05}
& Place
& $60.0 \pm 6.9$
& $88.0 \pm 4.6$
& {\boldmath$64.0 \pm 6.8$}
& {\boldmath$92.0 \pm 3.8$}
& {\boldmath$96.0 \pm 2.8$}
& {\boldmath$96.0 \pm 2.8$}
& $78.0 \pm 5.9$
& $48.0 \pm 7.1$
& \multirow{-2}{*}{\textbf{44.6}}
\\
\Xhline{0.8pt}

\end{tabular}%
}

\arrayrulecolor{black}
\endgroup
\vspace{-5mm}
\end{table*}

\paragraph{Skill- and object-level analysis.}
Performance varies substantially across both manipulation skills and
object categories. Place is consistently easier than Pick, making
reliable aerial object acquisition the primary bottleneck. Success
rates also differ markedly across objects, reflecting variations in
geometry, graspability, and manipulation difficulty. These differences
demonstrate that the benchmark captures meaningful variation across
models, skills, and objects rather than reducing evaluation to a single
aggregate success rate.

\section{Conclusion and Limitations}
We presented \methodname{}, a GPU-parallel benchmark for scalable aerial manipulation data generation and policy evaluation. The framework integrates payload-aware flight control, automated demonstration generation, and event-based trajectory annotation, providing more than 80K demonstrations spanning basic skills and long-horizon tasks that combine navigation and manipulation. Our experiments show that BC efficiently reproduces expert behavior, while PPO discovers more diverse and faster successful trajectories at a higher training cost. The evaluated pretrained VLA models achieve higher mean success rates than the visuomotor imitation baselines, although reliable aerial picking remains challenging. Our evaluation is currently limited to simulation, and sim-to-real transfer remains unvalidated. Future work will validate the framework on physical aerial manipulators and expand its coverage of platforms, environments, and long-horizon tasks.


\section*{Reproducibility Statement}

We open source all code for environments, training, evaluation, and data generation, and we release our dataset for public use.

\section*{AI Use Statement}
We used generative AI tools to polish the writing and improve the readability of the manuscript. The authors reviewed and verified all AI-assisted content and take full responsibility for the final manuscript.



\bibliography{iclr2027_conference}

@inproceedings{habitat2.0,
  author       = {Andrew Szot and
                  Alexander Clegg and
                  Eric Undersander and
                  Erik Wijmans and
                  Yili Zhao and
                  John M. Turner and
                  Noah Maestre and
                  Mustafa Mukadam and
                  Devendra Singh Chaplot and
                  Oleksandr Maksymets and
                  Aaron Gokaslan and
                  Vladimir Vondrus and
                  Sameer Dharur and
                  Franziska Meier and
                  Wojciech Galuba and
                  Angel X. Chang and
                  Zsolt Kira and
                  Vladlen Koltun and
                  Jitendra Malik and
                  Manolis Savva and
                  Dhruv Batra},
  editor       = {Marc'Aurelio Ranzato and
                  Alina Beygelzimer and
                  Yann N. Dauphin and
                  Percy Liang and
                  Jennifer Wortman Vaughan},
  title        = {Habitat 2.0: Training Home Assistants to Rearrange their Habitat},
  booktitle    = {Advances in Neural Information Processing Systems 34: Annual Conference
                  on Neural Information Processing Systems 2021, NeurIPS 2021, December
                  6-14, 2021, virtual},
  pages        = {251--266},
  year         = {2021}
}

@article{taomaniskill3,
  title={ManiSkill3: {GPU} Parallelized Robotics Simulation and Rendering for Generalizable Embodied AI},
  author={Stone Tao and Fanbo Xiang and Arth Shukla and Yuzhe Qin and Xander Hinrichsen and Xiaodi Yuan and Chen Bao and Xinsong Lin and Yulin Liu and Tse-kai Chan and Yuan Gao and Xuanlin Li and Tongzhou Mu and Nan Xiao and Arnav Gurha and Zhiao Huang and Roberto Calandra and Rui Chen and Shan Luo and Hao Su},
  journal = {arXiv preprint arXiv:2410.00425},
  year={2024},
}

@inproceedings{maniskillhab,
  title     = {{ManiSkill-HAB}: A Benchmark for Low-Level Manipulation in Home Rearrangement Tasks},
  author    = {Arth Shukla and Stone Tao and Hao Su},
  booktitle = {The Thirteenth International Conference on Learning Representations},
  year      = {2025}
}

@article{Air-vla,
  title={Air-vla: Vision-language-action systems for aerial manipulation},
  author={Sun, Jianli and Tian, Bin and Zhang, Qiyao and Li, Chengxiang and Song, Zihan and Cui, Zhiyong and Lv, Yisheng and Tian, Yonglin},
  journal={arXiv preprint arXiv:2601.21602},
  year={2026}
}

@article{AIR-VLA+,
  title={AIR-VLA+: Decoupling Movement and Manipulation via Cascaded Dual-Action Decoders with Asymmetric MoE for Aerial Robots},
  author={Sun, Jianli and Tian, Bin and Zhang, Qiyao and Liu, Zijian and Wang, Yutong and Cui, Zhiyong and Li, Bai and Lv, Yisheng and Tian, Yonglin},
  journal={arXiv preprint arXiv:2606.12859},
  year={2026}
}

@article{AIR-pi,
  title={$\pi$, but make it fly: Physics-guided transfer of VLA models to aerial manipulation},
  author={Tucker, Johnathan and Liu, Denis and Swann, Aiden and Ren, Allen and Yu, Javier and Sun, Jiankai and Kim, Brandon and McGranahan, Lachlain and Vuong, Quan and Schwager, Mac},
  journal={arXiv preprint arXiv:2603.25038},
  year={2026}
}

@article{LIBERO,
  title={Libero: Benchmarking knowledge transfer for lifelong robot learning},
  author={Liu, Bo and Zhu, Yifeng and Gao, Chongkai and Feng, Yihao and Liu, Qiang and Zhu, Yuke and Stone, Peter},
  journal={Advances in Neural Information Processing Systems},
  volume={36},
  pages={44776--44791},
  year={2023}
}

@inproceedings{li2023behavior,
  title={Behavior-1k: A benchmark for embodied ai with 1,000 everyday activities and realistic simulation},
  author={Li, Chengshu and Zhang, Ruohan and Wong, Josiah and Gokmen, Cem and Srivastava, Sanjana and Mart{\'\i}n-Mart{\'\i}n, Roberto and Wang, Chen and Levine, Gabrael and Lingelbach, Michael and Sun, Jiankai and others},
  booktitle={Conference on Robot Learning},
  pages={80--93},
  year={2023},
  organization={PMLR}
}

@inproceedings{maniskill-hab,
  title={Maniskill-hab: A benchmark for low-level manipulation in home rearrangement tasks},
  author={Shukla, Arth and Tao, Stone and Su, Hao},
  booktitle={International Conference on Learning Representations},
  volume={2025},
  pages={15288--15317},
  year={2025}
}

@inproceedings{nasiriany2026robocasa365,
  title={Robocasa365: A large-scale simulation framework for training and benchmarking generalist robots},
  author={Nasiriany, Soroush and Nasiriany, Sep and Maddukuri, Abhiram and Zhu, Yuke},
  booktitle={International Conference on Learning Representations},
  volume={2026},
  pages={98643--98667},
  year={2026}
}

@article{geng2025roboverse,
  title={Roboverse: Towards a unified platform, dataset and benchmark for scalable and generalizable robot learning},
  author={Geng, Haoran and Wang, Feishi and Wei, Songlin and Li, Yuyang and Wang, Bangjun and An, Boshi and Cheng, Charlie Tianyue and Lou, Haozhe and Li, Peihao and Wang, Yen-Jen and others},
  journal={arXiv preprint arXiv:2504.18904},
  year={2025}
}

@article{chen2025robotwin,
  title={Robotwin 2.0: A scalable data generator and benchmark with strong domain randomization for robust bimanual robotic manipulation},
  author={Chen, Tianxing and Chen, Zanxin and Chen, Baijun and Cai, Zijian and Liu, Yibin and Li, Zixuan and Liang, Qiwei and Lin, Xianliang and Ge, Yiheng and Gu, Zhenyu and others},
  journal={arXiv preprint arXiv:2506.18088},
  year={2025}
}

@article{AM-Bench,
  title={AM-Bench: A Modular Simulation Suite and Benchmark for Aerial Manipulation Policy Learning},
  author={Wang, Yutong and Lee, Dongjae and Guo, Xiaofeng and Zhan, Yuanzhu and Jiang, Yufei and Saravanan, Bavin and Cao, Muqing and Xie, Jia and Mao, Chenyang and Scherer, Sebastian and others},
  journal={arXiv preprint arXiv:2609.00641},
  year={2026}
}

@inproceedings{mehboob2026dronevla,
  title={DroneVLA: VLA-Based Aerial Manipulation},
  author={Mehboob, Fawad and James, MoniJesu Wonders and Habel, Amir Atef and Sam, Jeffrin and Altamirano Cabrera, Miguel and Tsetserukou, Dzmitry},
  booktitle={Companion Proceedings of the 21st ACM/IEEE International Conference on Human-Robot Interaction},
  pages={1135--1139},
  year={2026}
}

@article{mandlekar2023mimicgen,
  title={Mimicgen: A data generation system for scalable robot learning using human demonstrations},
  author={Mandlekar, Ajay and Nasiriany, Soroush and Wen, Bowen and Akinola, Iretiayo and Narang, Yashraj and Fan, Linxi and Zhu, Yuke and Fox, Dieter},
  journal={arXiv preprint arXiv:2310.17596},
  year={2023}
}

@inproceedings{jiang2025dexmimicgen,
  title={Dexmimicgen: Automated data generation for bimanual dexterous manipulation via imitation learning},
  author={Jiang, Zhenyu and Xie, Yuqi and Lin, Kevin and Xu, Zhenjia and Wan, Weikang and Mandlekar, Ajay and Fan, Linxi Jim and Zhu, Yuke},
  booktitle={2025 IEEE International Conference on Robotics and Automation (ICRA)},
  pages={16923--16930},
  year={2025},
  organization={IEEE}
}

@article{xue2025demogen,
  title={Demogen: Synthetic demonstration generation for data-efficient visuomotor policy learning},
  author={Xue, Zhengrong and Deng, Shuying and Chen, Zhenyang and Wang, Yixuan and Yuan, Zhecheng and Xu, Huazhe},
  journal={arXiv preprint arXiv:2502.16932},
  year={2025}
}

@article{garrett2024skillmimicgen,
  title={Skillmimicgen: Automated demonstration generation for efficient skill learning and deployment},
  author={Garrett, Caelan and Mandlekar, Ajay and Wen, Bowen and Fox, Dieter},
  journal={arXiv preprint arXiv:2410.18907},
  year={2024}
}

@article{raghunandan2022lodestar,
  title={Lodestar: Supporting independent learning and rapid experimentation through data-driven analysis recommendations},
  author={Raghunandan, Deepthi and Cui, Zhe and Krishnan, Kartik and Tirfe, Segen and Shi, Shenzhi and Shrestha, Tejaswi Darshan and Battle, Leilani and Elmqvist, Niklas},
  journal={arXiv preprint arXiv:2204.07876},
  year={2022}
}

@inproceedings{wang2025towards,
  title={Towards realistic uav vision-language navigation: Platform, benchmark, and methodology},
  author={Wang, Xiangyu and Yang, Donglin and Wang, Ziqin and Kwan, Hohin and Chen, Jinyu and Li, Hongsheng and Liao, Yue and Liu, Si and others},
  booktitle={International Conference on Learning Representations},
  volume={2025},
  pages={7292--7310},
  year={2025}
}

@article{wang2026uav,
  title={Uav-flow colosseo: A real-world benchmark for flying-on-a-word uav imitation learning},
  author={Wang, Xiangyu and Yang, Donglin and Liao, Yue and Zheng, Wenhao and Dai, Bin and Li, Hongsheng and Liu, Si and others},
  journal={Advances in Neural Information Processing Systems},
  volume={38},
  year={2026}
}

@article{DP,
  title={Diffusion policy: Visuomotor policy learning via action diffusion},
  author={Chi, Cheng and Xu, Zhenjia and Feng, Siyuan and Cousineau, Eric and Du, Yilun and Burchfiel, Benjamin and Tedrake, Russ and Song, Shuran},
  journal={The International Journal of Robotics Research},
  volume={44},
  number={10-11},
  pages={1684--1704},
  year={2025},
  publisher={Sage Publications Sage UK: London, England}
}

@article{ACT,
  title={Learning fine-grained bimanual manipulation with low-cost hardware},
  author={Zhao, Tony Z and Kumar, Vikash and Levine, Sergey and Finn, Chelsea},
  journal={arXiv preprint arXiv:2304.13705},
  year={2023}
}

@article{pi0.5,
  title={$\pi_{0.5}$: a Vision-Language-Action Model with Open-World Generalization},
  author={Intelligence, Physical and Black, Kevin and Brown, Noah and Darpinian, James and Dhabalia, Karan and Driess, Danny and Esmail, Adnan and Equi, Michael and Finn, Chelsea and Fusai, Niccolo and others},
  journal={arXiv preprint arXiv:2504.16054},
  year={2025}
}

@article{pi0,
  title={$\pi_{0}$: A Vision-Language-Action Flow Model for General Robot Control},
  author={Black, Kevin and Brown, Noah and Driess, Danny and Esmail, Adnan and Equi, Michael and Finn, Chelsea and Fusai, Niccolo and Groom, Lachy and Hausman, Karol and Ichter, Brian and others},
  journal={arXiv preprint arXiv:2410.24164},
  year={2024}
}

@article{Air-UMI,
  title={Umi-on-air: Embodiment-aware guidance for embodiment-agnostic visuomotor policies},
  author={Gupta, Harsh and Guo, Xiaofeng and Ha, Huy and Pan, Chuer and Cao, Muqing and Lee, Dongjae and Scherer, Sebastian and Song, Shuran and Shi, Guanya},
  journal={arXiv preprint arXiv:2510.02614},
  year={2025}
}

@article{UMI,
  title={Universal manipulation interface: In-the-wild robot teaching without in-the-wild robots},
  author={Chi, Cheng and Xu, Zhenjia and Pan, Chuer and Cousineau, Eric and Burchfiel, Benjamin and Feng, Siyuan and Tedrake, Russ and Song, Shuran},
  journal={arXiv preprint arXiv:2402.10329},
  year={2024}
}
\bibliographystyle{iclr2027_conference}

\clearpage
\appendix
\section{Appendix}
\subsection{Task Examples and Qualitative Rollouts}
\label{app:task_examples}
\begin{figure}[t!]
    \centering
    \includegraphics[width=0.9\linewidth]{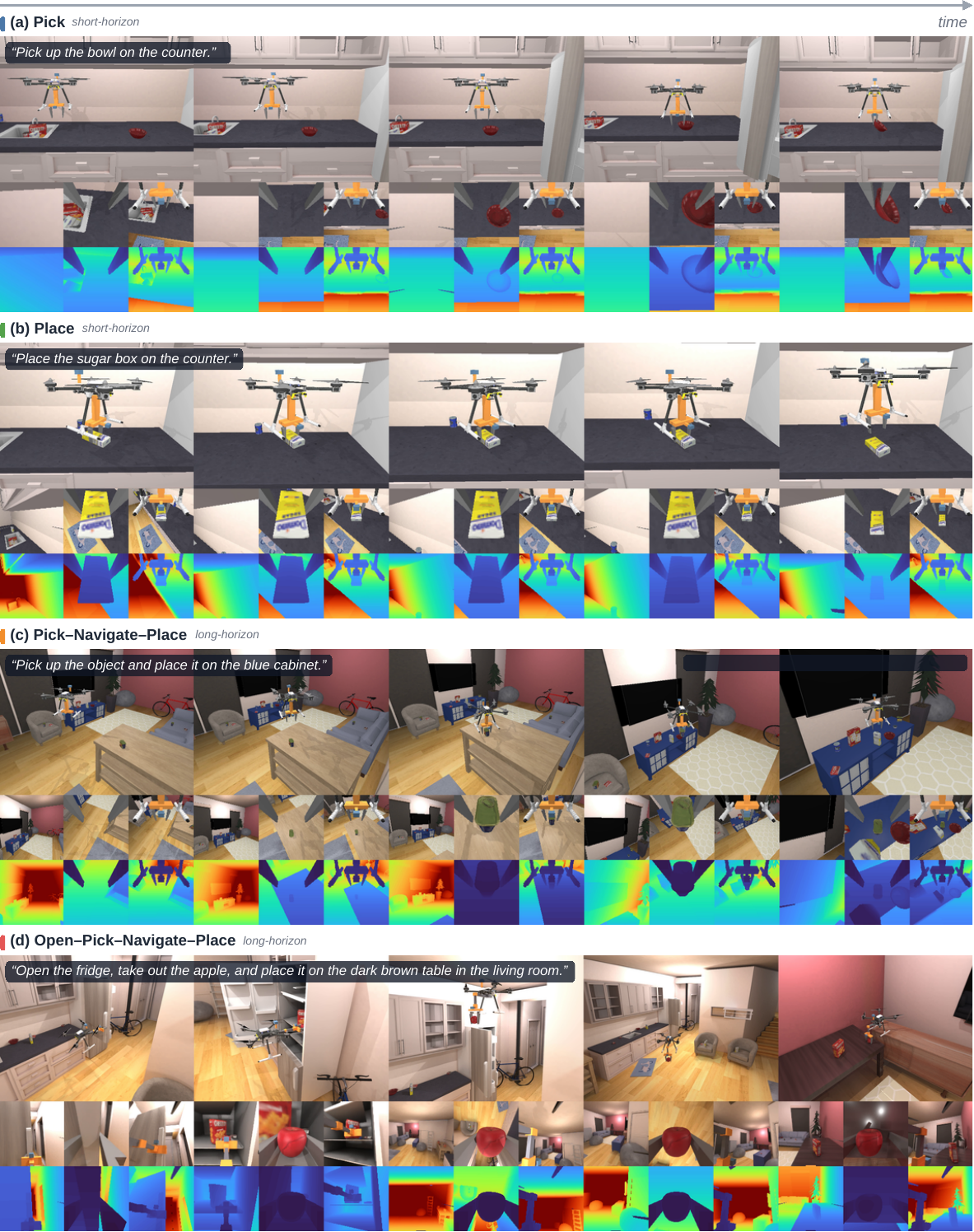}
    \vspace{-2mm}
    \caption{
     Representative qualitative rollouts in \methodname{}:
        (a) short-horizon \textsc{Pick},
        (b) short-horizon \textsc{Place},
        (c) long-horizon \textsc{Pick--Navigate--Place}, and
        (d) long-horizon \textsc{Open--Pick--Navigate--Place}.
        Each sequence is paired with a language instruction and
        progresses from left to right in time.
        For each task, the top row shows third-person views,
        while the middle and bottom rows show onboard RGB and
        depth observations, respectively.
        Third-person views are used only for visualization and
        are not provided to the evaluated policies. \textbf{All demonstration videos are available in the supplementary materials.}
    }
    \label{fig:demos}
    \vspace{-7mm}
\end{figure}

Figure~\ref{fig:demos} presents representative rollouts of four
aerial manipulation tasks in \methodname{}.
Short-horizon \textsc{Pick} and \textsc{Place} focus on local
object interaction, while long-horizon
\textsc{Pick--Navigate--Place} requires aerial transport between
spatially separated source and destination regions.
\textsc{Open--Pick--Navigate--Place} further incorporates
articulated-object interaction: the robot must open a refrigerator
before retrieving, transporting, and placing the target object.
Additional demonstration videos are provided in the supplementary
material.

\subsection{Observation Space}
\label{app:observation}

\begin{figure}[t!]
    \centering
    \includegraphics[width=\linewidth]{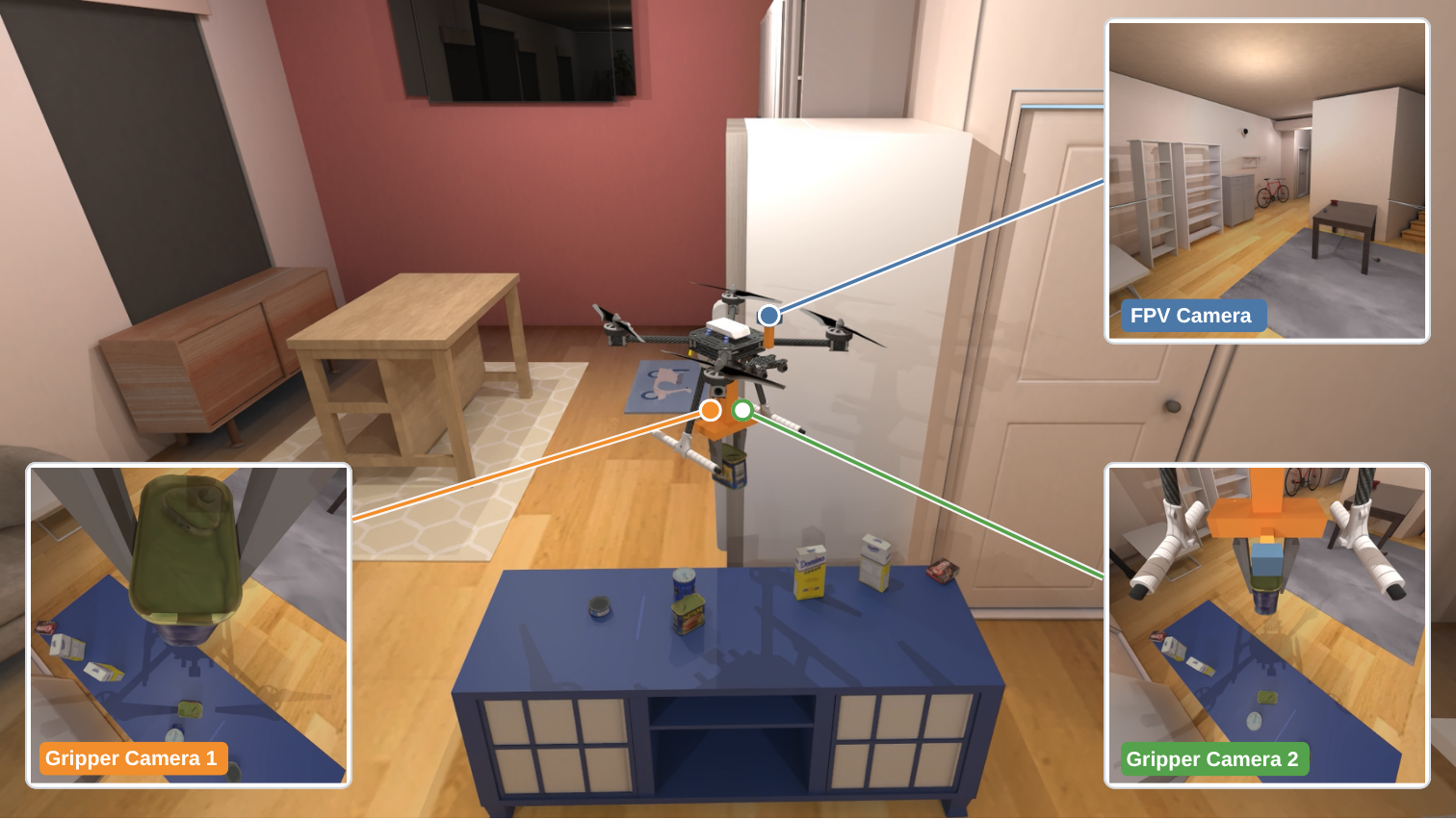}
    \caption{
    Observation configuration of \methodname{}.
    The aerial platform is equipped with one forward-facing FPV camera
    and two gripper-mounted cameras, providing complementary global and
    local views for navigation and manipulation. The central third-person
    view illustrates the camera placement and is not provided to the
    evaluated policy.
    }
    \label{fig:observation}
\end{figure}

Our observation space combines multi-view visual inputs with robot
proprioception. As shown in Figure~\ref{fig:observation}, the aerial
platform is equipped with three body-mounted RGB-D cameras: one
forward-facing FPV camera for global scene perception and navigation,
and two gripper-mounted cameras that provide complementary views of
the local manipulation workspace. The onboard RGB-D streams are recorded at $128\times128$ resolution.
In addition to visual observations, the policy receives a
19-dimensional proprioceptive state containing gripper states, UAV
position and orientation, and linear and angular velocities.
Language-conditioned policies additionally receive a natural-language
task instruction. The third-person view shown in Figure~\ref{fig:observation} is used
only to illustrate the camera configuration and is not provided to
the evaluated policy.

\subsection{GPU-Parallel Data Collection and Inference}
\begin{figure}[t]
  \centering
  \includegraphics[width=0.75\linewidth]{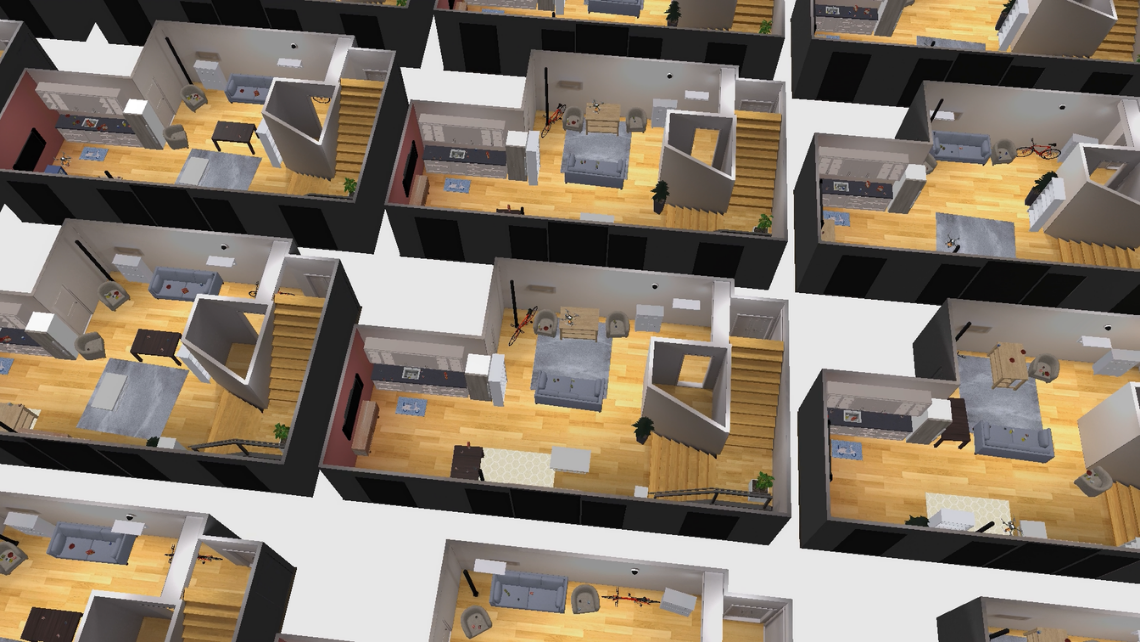}
    \caption{
        GPU-parallel data collection and inference in AeroManip-VLA.
    Reinforcement learning policies automatically generate diverse
    aerial manipulation demonstrations across parallel simulation
    environments without human teleoperation, while batched VLA
    inference enables efficient, large-scale evaluation.
    }
    \label{fig:sim_gpu}
\end{figure}

Unlike AIR-VLA, which relies on human teleoperation to collect
demonstrations, our framework uses reinforcement learning policies
to automatically generate aerial manipulation trajectories across
parallel simulation environments. This enables scalable collection
of large, diverse datasets without manual teleoperation. Beyond data collection, our framework supports batched policy
inference across parallel environments, enabling efficient,
large-scale evaluation of Vision-Language-Action models. Together,
automated data generation and parallel inference provide a
scalable foundation for aerial manipulation research. The simulator runs on a range of NVIDIA GPUs, including H200,
A100, and GeForce RTX 5090, supporting deployment on both
data-center and consumer hardware.

\subsection{Ray Tracing and Visual Fidelity}
\begin{figure}[h!]
    \centering
    \includegraphics[width=\textwidth]{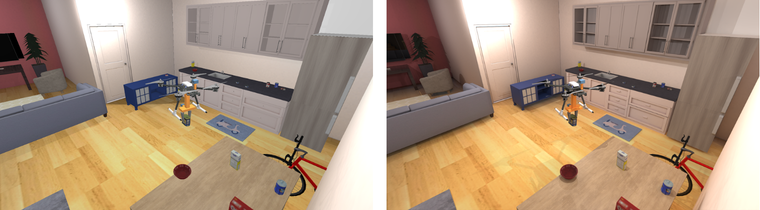}
    \caption{
        Qualitative comparison of \methodname ~with ray tracing
        disabled (left) and enabled (right).
        Both views are rendered directly by the simulator,
        illustrating differences in lighting, shadows, and
        surface shading.
    }
    \label{fig:raytracingcomp}
\end{figure}

\methodname supports optional ray-traced rendering to enhance the
visual fidelity of aerial manipulation scenes.
Figure~\ref{fig:raytracingcomp} compares the same scene with ray
tracing disabled and enabled, highlighting differences in
illumination, shadows, and material appearance.
This option complements non-ray-traced rendering with visually
richer observations for demonstration collection and aerial
VLA evaluation.

\subsection{Additional Experiments}
\subsubsection{Simulation Throughput and Scalability}
To assess scalability for visual data collection and on-policy RL,
we measure aggregate
simulation-and-rendering throughput and GPU memory usage as the
number of parallel environments $N$ increases.
Each environment contains a full ReplicaCAD apartment,
TidyHouse objects, and our aerial manipulator, initialized
$0.5$--$1$\,m from its target object.
Every control step renders three onboard $128\times128$ RGB-D
cameras, with physics running at $240$\,Hz and control at $20$\,Hz.
After an untimed reset, we measure the wall-clock time $t$ for
$200$ control steps of seeded random velocity and gripper
commands, reporting samples per second
($\mathrm{SPS}=200N/t$) and total GPU memory usage.
We benchmark AIR-VLA~\citep{Air-vla} in Isaac Sim on the same RTX 5090
with its native evaluation settings: each environment renders four
$640\times480$ RGB policy cameras per control step, with physics at
$60$\,Hz and control at $20$\,Hz, and we time $250$ control steps of
open-loop demonstration replay after an untimed reset, computing SPS
and GPU memory in the same way.

\begin{figure}[h!]
    \centering
    \includegraphics[width=0.6\textwidth]
        {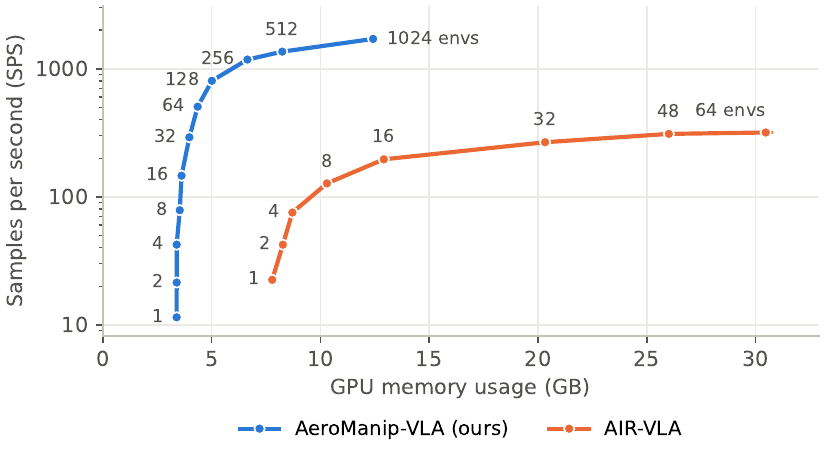}
    \caption{
        Interact benchmark comparing simulation-and-rendering
        throughput and GPU memory usage on an RTX 5090.
        Point labels indicate the number of parallel environments;
        per control step, each AeroManip-VLA environment renders three
$128\times128$ RGB-D cameras and each AIR-VLA environment renders
four $640\times480$ RGB cameras.
        Points show means over $10$ seeds, and error bars indicate
        $95\%$ confidence intervals.
        The vertical axis uses a logarithmic scale.
        At their highest-throughput tested configurations
        ($1024$ vs.\ $64$ environments), AeroManip-VLA achieves
        $5.4\times$ the throughput of AIR-VLA while using
        $41\%$ of its GPU memory.
    }
    \label{fig:interact_bench}
\end{figure}

Figure~\ref{fig:interact_bench} shows the results.
Among the tested configurations, AeroManip-VLA reaches
$1707$ SPS with $1024$ environments using $12.4$\,GB,
whereas AIR-VLA reaches its highest measured throughput of
$317$ SPS with $64$ environments using $30.5$\,GB.
At these respective configurations, AeroManip-VLA achieves
$5.4\times$ the throughput while using $41\%$ of the GPU memory.
At the same parallelism level of $N=64$, AeroManip-VLA achieves
approximately $1.6\times$ the throughput and uses $86\%$ less
GPU memory ($4.4$ vs.\ $30.5$\,GB).
AIR-VLA is faster at the tested configurations with $N\leq16$,
while the two methods have comparable throughput at $N=32$.
These results demonstrate the advantage of AeroManip-VLA
for highly parallel simulation and visual observation collection.

\subsubsection{Payload-Aware Flight Control}
\label{app:payload}

This appendix measures what the payload-aware switches of Sec.~\ref{sec:method:control} contribute. With payload-aware
control, three switches act on the carry. After three grasped control steps, the controller adds the object mass to the
feedforward ($\hat m \leftarrow m_0 + m_{\mathrm{obj}}$) and raises the attitude gains from $(k_R, k_\omega) = (2.5, 0.55)$
to $(12, 1.2)$. Once the support carries the object during placement, it restores $\hat m \leftarrow m_0$.
We compare this against the \emph{nominal} controller, which keeps $\hat m = m_0$ and the nominal gains throughout the carry.

\textbf{Why the nominal controller struggles with heavy objects.}
With $\hat m = m_0$, the object's weight must be carried by the integral term of Eq.~\eqref{eq:vel}. In steady hover this
requires $\xi_z^{\star} = g\,m_{\mathrm{obj}}/m_0$, which is 1.51--1.65\,m/s$^2$ for the three filled cans
(0.37--0.40\,kg), i.e.\ 75--83\% of the integrator's range of $\pm 2$\,m/s$^2$. This leaves little authority for climbing
and for rejecting disturbances. The integrator also fills slowly. While the drone pinches the object on its support but cannot
yet lift it, the vertical velocity error equals the commanded climb speed (0.15\,m/s during the expert's lift), so $\xi_z$ grows
by only 0.15\,m/s$^2$ per second. The object leaves the support once $3\,e_{v,z} + \xi_z \geq g\,m_{\mathrm{obj}}/m_0$,
which for the cans takes several seconds.

\textbf{Protocol.}
We run the GPU-batched expert on the same 126 TidyHouse instances (nine object classes,
14 instances each) with identical seeds and start states. The two conditions differ only in the payload-aware switches.
The stiff attitude gains used for release (placement stage 2 onward) belong to the release procedure of every object and
are active in both conditions.
The same 11 instances are rejected at initialization in both conditions, leaving 115 evaluated episodes per condition.
At every control step we log the airframe pose and angular velocity, the integrator state $\xi$, the rotor thrusts, and the
object's pose and angular velocity.
The \emph{carry} is the interval from the grasp handoff to the start of placement.
\emph{Lift-off} is the first time after the grasp at which the object has risen 5\,cm.
\emph{Tilt} is the angle between the airframe $z$-axis and vertical, and the \emph{tilt rate} is the magnitude of the
roll--pitch angular velocity.
\emph{Slip} is the object's height relative to the gripper's tool center point, measured from its value at the grasp.
Each condition is run once.

\begin{table}[h]
  \centering
  \small
  \caption{Outcomes of the expert on 115 TidyHouse episodes with and without payload-aware control (same instances,
  seeds and start states).}
  \label{tab:payload_outcomes}
  \begin{tabular}{@{}lrr@{}}
    \toprule
    Outcome & Payload-aware & Nominal \\
    \midrule
    Success                          & 106 (92.2\%) & 47 (40.9\%) \\
    Grasp lost                       & 3            & 33 \\
    Collision (during pick / transport) & 0 / 0     & 14 / 5 \\
    Timeout                          & 6            & 16 \\
    \bottomrule
  \end{tabular}
\end{table}

Table~\ref{tab:payload_outcomes} summarizes the outcomes. Without payload-aware control, success drops from 92.2\% to 40.9\%.
Most additional failures are grasps lost during transport (33) and collisions while the drone is still pinned at the
source (14). The nominal controller also times out 16 times, which never settles enough to
start the placement.
The nine remaining failures with payload-aware control are 3 lost grasps and 6 timeouts. Five of those timeouts are heavy
cans that do not settle at the end of the transport.

\begin{figure}[t]
  \centering
  \includegraphics[width=0.85\linewidth]{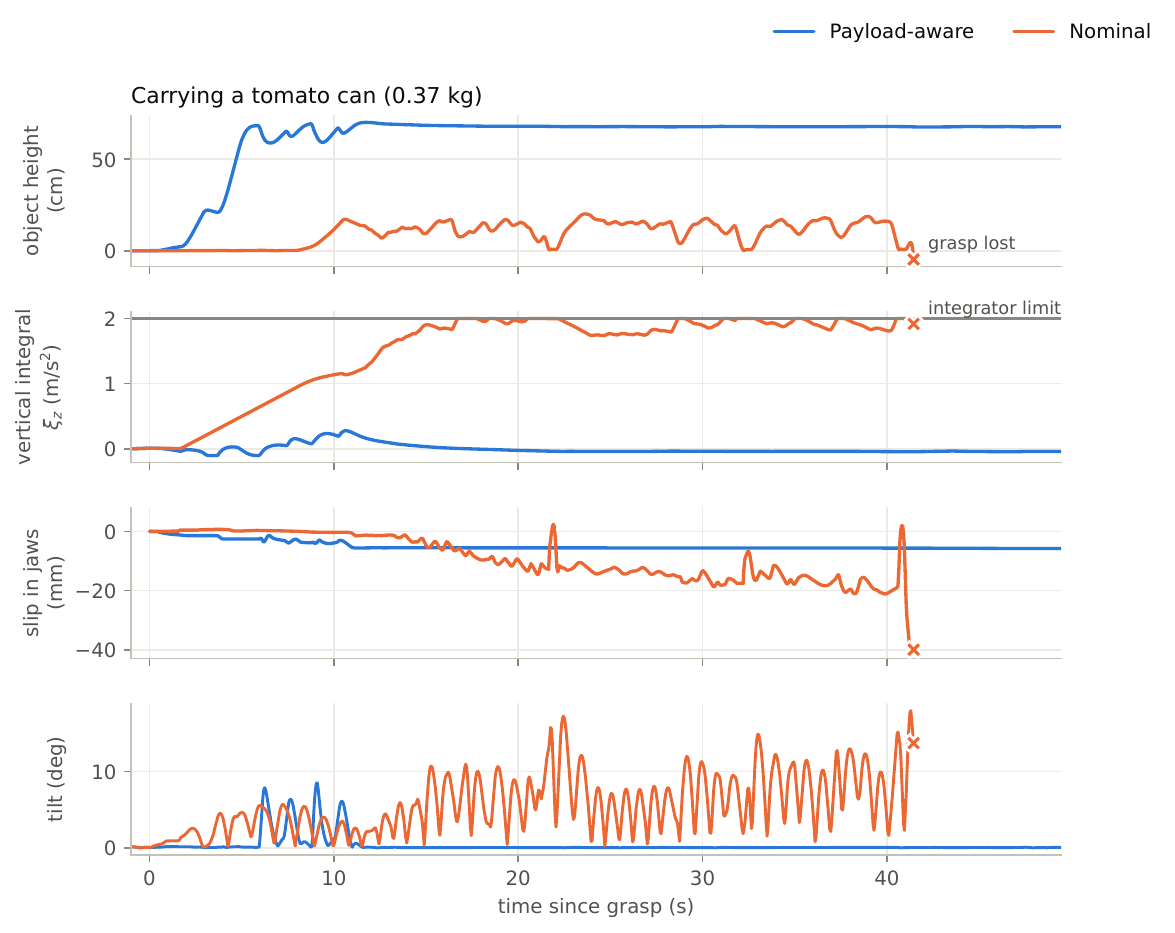}
  \caption{One tomato-can episode (0.37\,kg) with payload-aware control (blue) and the nominal controller (orange),
  from the same start state, aligned at the grasp. From top to bottom: object height above its initial rest height,
  vertical integrator state $\xi_z$ (the horizontal line marks the $\pm2$ limit), object slip in the jaws, and airframe tilt.
  $\times$ marks the loss of the grasp.}
  \label{fig:payload_timeseries}
\end{figure}

\textbf{A single carry (Fig.~\ref{fig:payload_timeseries}).}
We show one tomato-can instance chosen by a fixed rule, not by inspection. Among the 19 heavy-can instances that succeed
with payload-aware control and lose the grasp under the nominal controller, it is the one with the median nominal episode
length.
With payload-aware control, the object lifts off 2.1\,s after the grasp and reaches cruise height within 5\,s. $\xi_z$ stays
near zero, and once the transport has settled the tilt stays below $0.5^\circ$.
Under the nominal controller, the drone keeps pinching the can on its support while $\xi_z$ ramps up linearly; the can lifts
off only after 9.3\,s. $\xi_z$ reaches its limit at 16.7\,s and remains close to it for the rest of the carry.
The drone never reaches cruise height: the object stays 0--20\,cm above its support.
Meanwhile the airframe oscillates with tilts of up to $18^\circ$, and the can slides about 2\,cm down in the jaws until the
grasp is lost at 41.5\,s.

\begin{figure}[t]
  \centering
  \includegraphics[width=\linewidth]{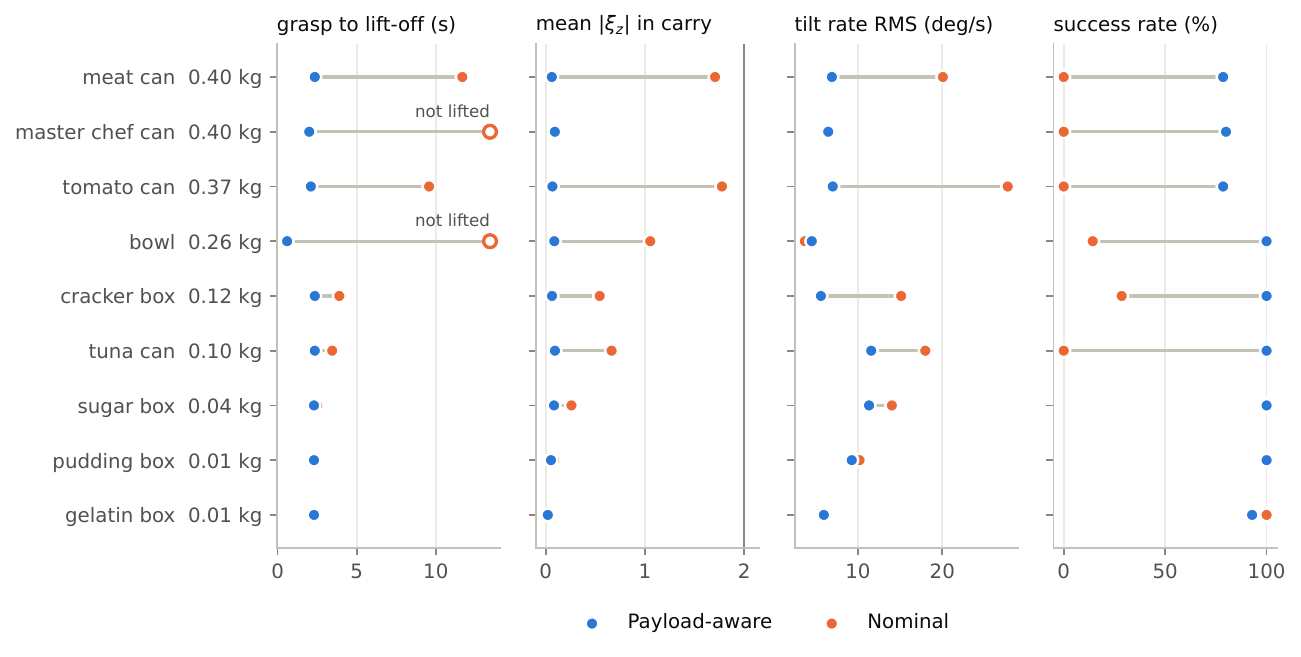}
  \caption{Per object class, sorted by mass, with the same instances in both conditions (TidyHouse, 14 instances per
  class, minus initialization rejections). Left to right: median time from grasp to lift-off (open marker: the median
  episode never lifts the object), mean $|\xi_z|$ during the carry (vertical line: integrator limit), RMS tilt rate during
  the carry, and success rate. Carry metrics are averaged over episodes that reach the carry, so they are missing where
  no nominal episode does (master chef can).}
  \label{fig:payload_objects}
\end{figure}

\textbf{Effect by object class (Fig.~\ref{fig:payload_objects}).}
The effect grows with payload mass.
With payload-aware control, the median lift-off time of every class is 0.6--2.4\,s, and the mean $|\xi_z|$ during the carry
stays below 0.1.
Under the nominal controller, lift-off takes 9.6\,s for the tomato can and 11.7\,s for the meat can. The master chef can is
never lifted, and the bowl is not lifted in four of seven episodes. The carry-time $|\xi_z|$ rises with mass up to 1.78
(tomato can), and the RMS tilt rate rises from 7.0 to 27.8\,deg/s (tomato can) and from 5.6 to 15.1\,deg/s (cracker box).
Success falls to 0\% for all three filled cans (from 79--80\%) and for the tuna can (from 100\%), to 29\% for the cracker
box and to 14\% for the bowl. The three lightest classes ($\le 0.04$\,kg: sugar, pudding and gelatin boxes) succeed in
93--100\% of episodes in both conditions.
The bowl is the one class whose tilt rate is not higher under the nominal controller (3.7 vs.\ 4.5\,deg/s). However, only
three nominal bowl episodes reach the carry at all.

\begin{figure}[t]
  \centering
  \includegraphics[width=\linewidth]{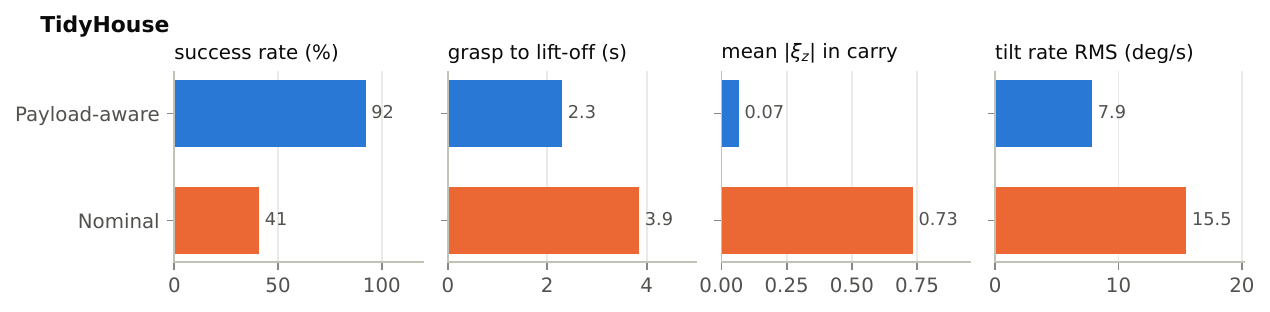}
  \caption{
    Overall effect of payload-aware control on TidyHouse.
    Compared with the nominal controller, payload-aware control achieves
    higher success rate, faster grasp-to-lift-off, lower mean vertical
    tracking error during transport, and lower RMS tilt rate.
    Values are aggregated over the evaluated episodes.
    }
  \label{fig:payload_ablation}
\end{figure}

\textbf{Overall effect of payload-aware control.}
Figure~\ref{fig:payload_ablation} summarizes the aggregate effect of
payload-aware control on TidyHouse. Compared with the nominal
controller, payload-aware control increases the success rate from
$41\%$ to $92\%$ and reduces the median grasp-to-lift-off time from
$3.9\,\mathrm{s}$ to $2.3\,\mathrm{s}$. During object transport, the
mean vertical tracking error decreases from $0.73$ to $0.07$, while
the RMS tilt rate is reduced from $15.5$ to $7.9\,\mathrm{deg/s}$.
Overall, payload-aware control improves both task completion and
flight stability during aerial manipulation.

\begin{figure}[t]
    \centering
    \includegraphics[width=.8\linewidth]
        {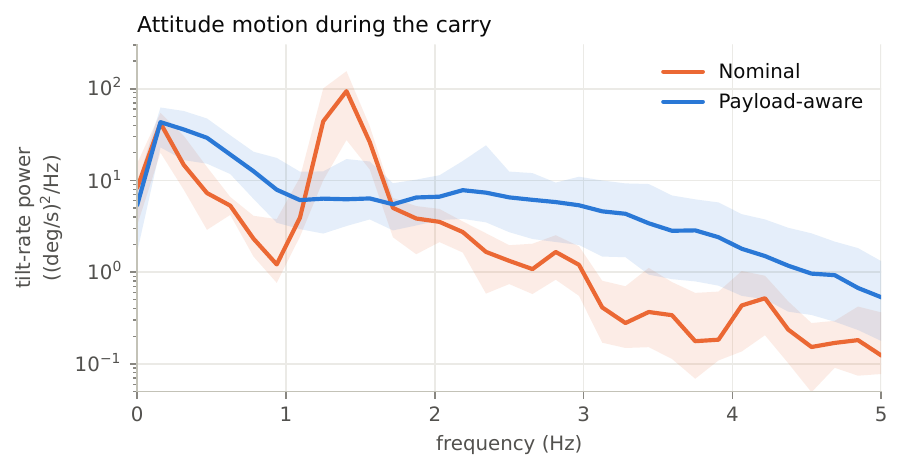}
    \caption{
        Power spectral density (PSD) of airframe tilt rate during
        TidyHouse carries lasting at least $6.4$\,s
        ($115$ payload-aware and $101$ nominal episodes).
        PSDs are estimated per episode using Welch's method with
        $6.4$\,s segments sampled at $20$\,Hz.
        Solid lines show pointwise medians across episodes;
        shaded bands indicate the 25th--75th percentiles.
        Payload-aware control attenuates the pronounced nominal
        peak near $1.4$\,Hz, while exhibiting higher PSD over
        $2$--$5$\,Hz.
    }
    \label{fig:payload_spectrum}
\end{figure}

\textbf{Carry oscillation (Fig.~\ref{fig:payload_spectrum}).}
The nominal controller exhibits a pronounced tilt-rate spectral
peak near $1.4$\,Hz. At this frequency, payload-aware control
reduces the median PSD from $94$ to $6.2$\,(deg/s)$^2$/Hz,
an approximately $15\times$ reduction.
Although the payload-aware spectrum is higher over $2$--$5$\,Hz,
the object-wise analysis in Fig.~\ref{fig:payload_objects}
shows lower RMS tilt rates for all object classes except the
bowl and gelatin box, where the absolute differences are below
$1$\,deg/s.
Together, these results support suppression of the pronounced
narrowband oscillation and reduced tilt-rate variation for
most object classes, rather than a uniform reduction in
spectral power across all frequencies.

\subsubsection{Failure Analysis}
\label{app:failure_analysis}

\paragraph{Motivation.}
Aggregate success rates do not reveal whether a policy fails during approach,
produces a weak grasp that fails later, or releases an object in an unstable
state. We therefore analyze failures at the object, mechanism, and skill-transition
levels. Besides characterizing the benchmark beyond a single success metric,
this analysis exposes actionable failure signatures for dataset users. In
particular, it can guide class-balanced sampling and augmentation, contact-aware
approach control, grasp-quality and handoff objectives, and stability-aware
release strategies. The transition analysis is important because a policy may
perform well from expert-generated initial states yet degrade when it receives
states produced by another learned skill.

\paragraph{Protocol.}
We evaluate four held-out starts for each of 111 \textsc{TidyHouse} and 110
\textsc{PrepareGroceries} instances, giving 444 and 440 episodes, respectively.
In the chained evaluation, the pick policy acts for at most 400 control steps,
the same expert controller executes the planned transport route for at most
1200 steps, and the place policy takes control 0.5--1.0\,m from the target for
at most 1000 steps. Because transport is shared across all chains, differences
in its outcome primarily reflect the state produced by the preceding pick.
During transport, a grasp is considered maintained while both fingers exert
more than 0.1\,N on the object and the tool center point remains within 6\,cm
of the planned grasp point; four consecutive violations terminate the transport.
For the mechanism-level analyses, a diagnostic rerun additionally records the
speed at first finger contact, object rotation at the pick-to-transport handoff,
and the condition under which the grasp is lost. The rerun covers all four starts
per instance for PPO and one \textsc{TidyHouse} or two
\textsc{PrepareGroceries} starts per instance for the expert. Each PPO skill is
represented by one training run, so the results diagnose these policies rather
than estimate variation across random seeds.

\begin{figure}[t]
  \centering
  \includegraphics[width=\linewidth]{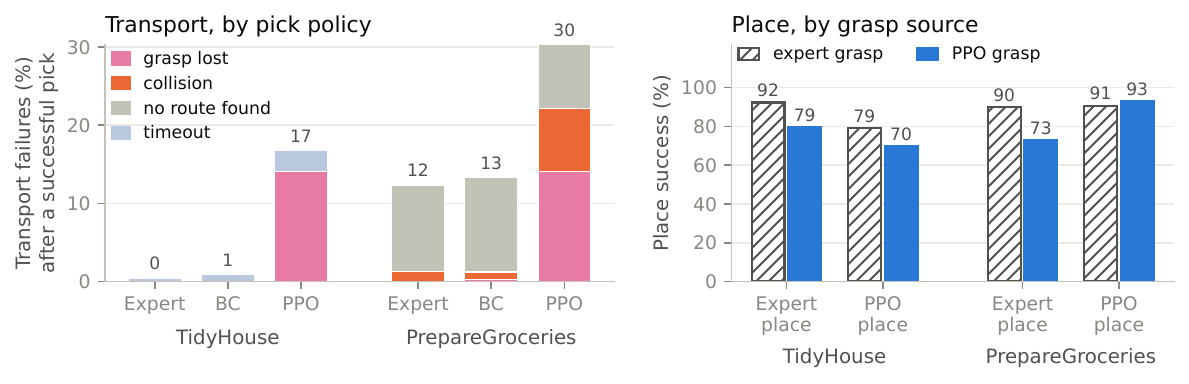}
  \caption{
    Error propagation across skill boundaries.
    (a) Failure rate of the shared expert transport after a successful pick,
    grouped by the policy that produced the grasp and by failure cause.
    (b) Place success conditioned on successful transport, separated by place
    policy and grasp source. For \textsc{PrepareGroceries}, the expert-grasp
    result for PPO place is taken from the held-out place evaluation, whose
    initial states are generated by expert transport.
  }
  \label{fig:fail_inherit}
\end{figure}

\paragraph{Errors propagate across learned skills (Fig.~\ref{fig:fail_inherit}).}
After a successful expert or BC pick, transport fails in at most 0.9\% of
\textsc{TidyHouse} episodes and in 12--13\% of \textsc{PrepareGroceries}
episodes, where most failures are caused by the route planner. With a PPO grasp,
the same transport fails in 16.7\% and 30.3\% of episodes, respectively;
grasp loss alone accounts for 14.1\% in both tasks. The grasp source also affects
the next skill: the expert place policy drops from 92.3\% to 79.4\% in
\textsc{TidyHouse} and from 89.9\% to 72.9\% in
\textsc{PrepareGroceries} when expert grasps are replaced by PPO grasps.
The corresponding PPO-place rates are 79.2\% versus 69.9\% in
\textsc{TidyHouse}, while they remain similar in \textsc{PrepareGroceries}
(90.6\% versus 93.0\%). These results show that evaluating each skill only from
expert-generated states can overestimate end-to-end performance; training and
evaluation should also include policy-induced handoff states.

\begin{figure}[t]
  \centering
  \includegraphics[width=\linewidth]{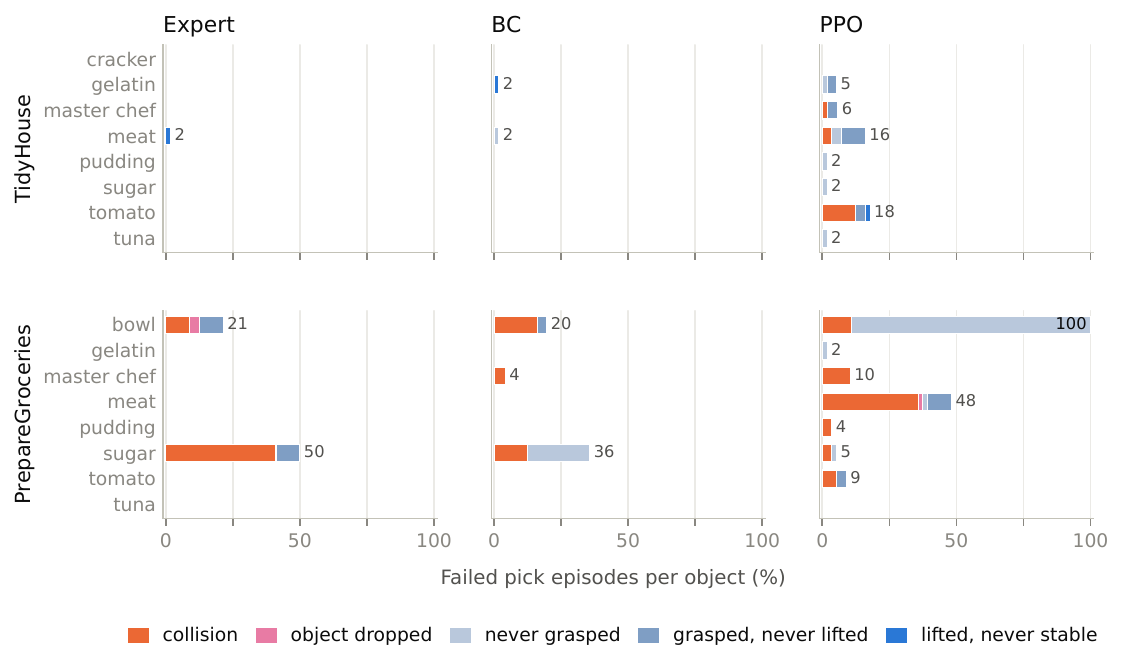}
  \caption{
    Pick-failure rate for each object class on held-out starts, grouped by policy
    and failure mode. Percentages are computed within each class. Timeout cases
    are categorized by the last stage reached: never grasped, grasped but never
    lifted, or lifted but never stabilized.
  }
  \label{fig:fail_pick_objects}
\end{figure}

\paragraph{Pick failures are object dependent (Fig.~\ref{fig:fail_pick_objects}).}
Failures are concentrated in a small number of classes rather than distributed
uniformly. In \textsc{PrepareGroceries}, PPO fails on every bowl episode and on
48\% of potted-meat episodes, whereas no other class exceeds 11\%. The expert
and BC also struggle with bowls (21\% and 20\%) and sugar boxes (50\% and 36\%),
but for different failure modes. In \textsc{TidyHouse}, PPO's largest failure
rates occur for tomato cans (18\%) and potted-meat cans (16\%). This concentration
motivates per-class reporting and object-balanced sampling; aggregate success
alone would hide both rare hard classes and policy-specific weaknesses.

\begin{figure}[t]
  \centering
  \includegraphics[width=\linewidth]{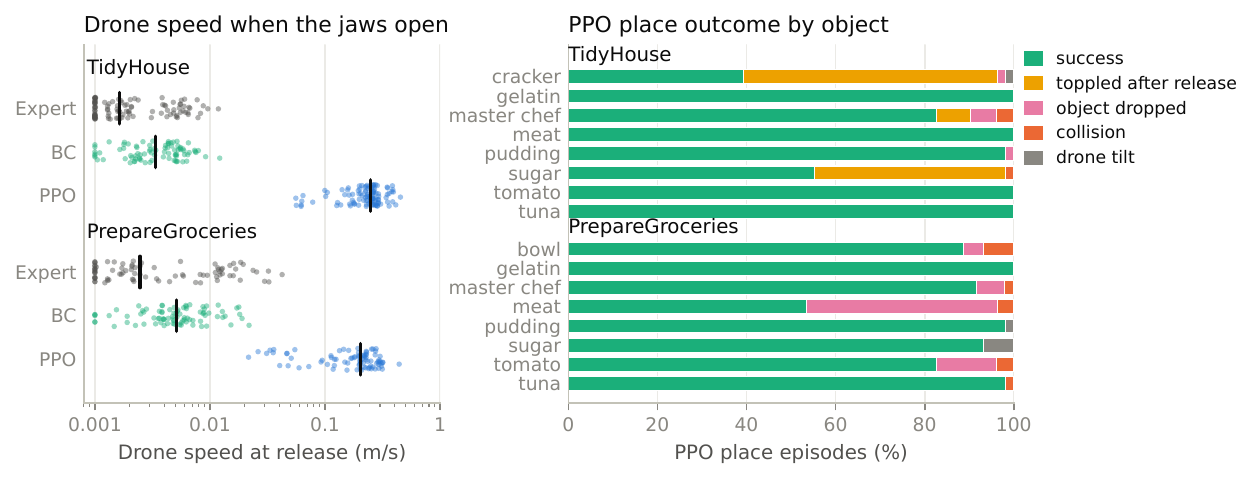}
  \caption{
    Placement behavior.
    (a) Drone speed when the jaws open on the first held-out start of each
    instance; each dot is one episode and the vertical bar is the median.
    (b) PPO place outcomes by object class over all held-out place starts.
    ``Toppled after release'' denotes an object that was released and followed
    by retreat but did not come to rest upright inside the goal region.
  }
  \label{fig:fail_place}
\end{figure}

\paragraph{Moving releases reduce placement stability (Fig.~\ref{fig:fail_place}).}
The expert and BC nearly stop before opening the jaws, with median release speeds
of 0.0016--0.0025\,m/s and 0.0034--0.0051\,m/s, respectively. PPO instead
releases at 0.20--0.25\,m/s. This behavior is often sufficient for squat objects
but is brittle for tall or heavy objects: in \textsc{TidyHouse}, cracker and
sugar boxes topple after release in 57\% and 43\% of episodes, while in
\textsc{PrepareGroceries}, potted-meat and tomato cans are dropped in 43\% and
13\% of episodes. A stop-before-release constraint or an explicit post-release
stability objective would directly target this failure mode.

\begin{figure}[t]
  \centering
  \includegraphics[width=\linewidth]{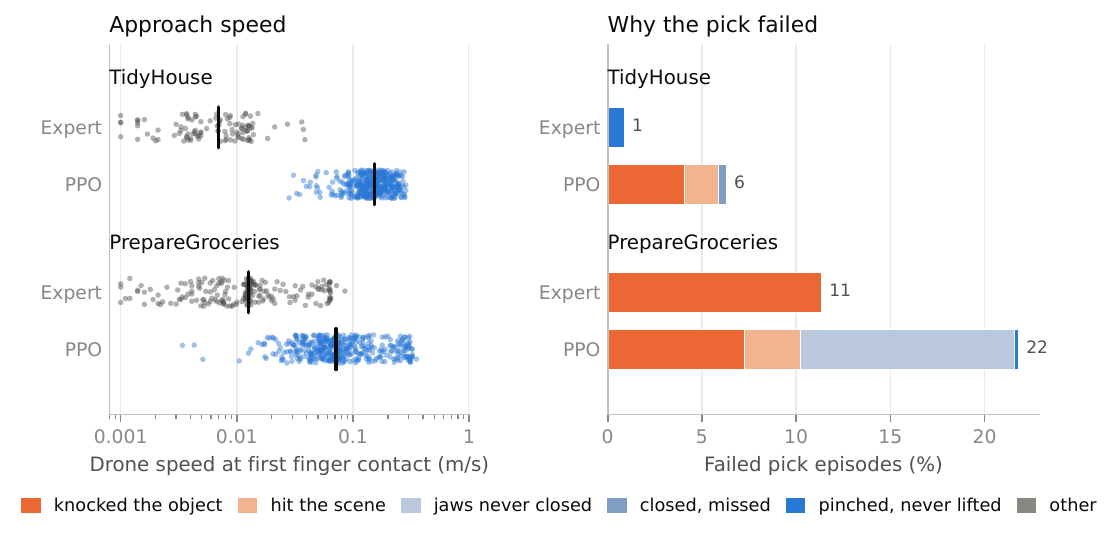}
  \caption{
    Pick mechanisms in the diagnostic rerun.
    (a) Drone speed at first finger contact; the vertical bar is the median.
    (b) Failed pick episodes grouped by mechanism. ``Knocked the object'' means
    that the object moved more than 3\,cm or rotated more than 20$^\circ$ before
    being grasped; ``hit the scene'' denotes a collision while the object was
    undisturbed.
  }
  \label{fig:fail_pick_mech}
\end{figure}

\paragraph{Fast approaches and missed closure dominate PPO pick failures
(Fig.~\ref{fig:fail_pick_mech}).}
At first finger contact, PPO's median speed is 0.152\,m/s in
\textsc{TidyHouse} and 0.072\,m/s in \textsc{PrepareGroceries}, compared with
0.0070\,m/s and 0.0126\,m/s for the expert. Consequently, PPO frequently pushes
or tips the object before closing the jaws, accounting for 4.1\% and 6.4\% of
all episodes. A separate \textsc{PrepareGroceries} failure occurs in 11.4\% of
episodes: PPO reaches the planned grasp pose but never closes the gripper,
including 48 bowl episodes. The expert can grasp the same bowls, indicating an
exploration or action-selection failure rather than a geometric impossibility.
These observations motivate approach deceleration, contact-aware control, and
explicit supervision or exploration for gripper closure and rim grasps.

\begin{figure}[t]
  \centering
  \includegraphics[width=\linewidth]{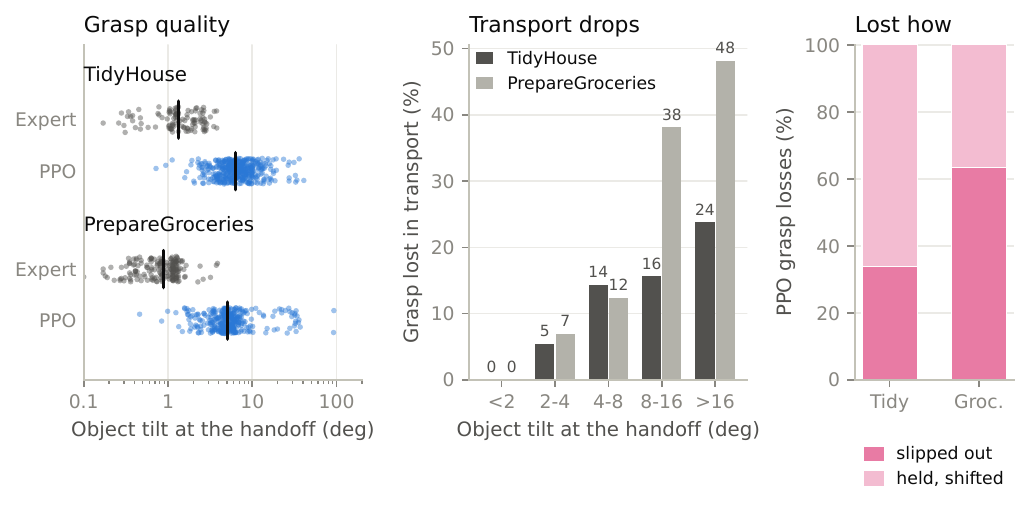}
  \caption{
    Grasp quality and transport robustness.
    (a) Object rotation relative to its pre-pick orientation at the
    pick-to-transport handoff; the vertical bar is the median.
    (b) Grasp-loss rate during expert transport as a function of handoff
    rotation, pooling expert and PPO picks.
    (c) Mechanisms of PPO grasp loss. Bowls are excluded because their planned
    rim grasp rotates them by design.
  }
  \label{fig:fail_grasp}
\end{figure}

\paragraph{Handoff rotation predicts transport failure
(Fig.~\ref{fig:fail_grasp}).}
Objects picked by the expert rotate by a median of 1.3$^\circ$ in
\textsc{TidyHouse} and 0.9$^\circ$ in \textsc{PrepareGroceries}; the PPO
medians rise to 6.4$^\circ$ and 5.1$^\circ$. No grasp with less than
2$^\circ$ of handoff rotation is lost, whereas the loss rate reaches 24\% in
\textsc{TidyHouse} and 48\% in \textsc{PrepareGroceries} above 16$^\circ$.
The mechanism differs across tasks: 63\% of PPO losses in
\textsc{PrepareGroceries} are physical slips, while 66\% in
\textsc{TidyHouse} remain pinched but move beyond the 6\,cm tolerance around
the planned grasp point. Handoff orientation and grasp-point deviation are
therefore useful training targets and diagnostics, not merely auxiliary metrics.

\subsection {Experimental Implementation}
\begin{figure}[t]
    \centering
    \includegraphics[width=\linewidth]{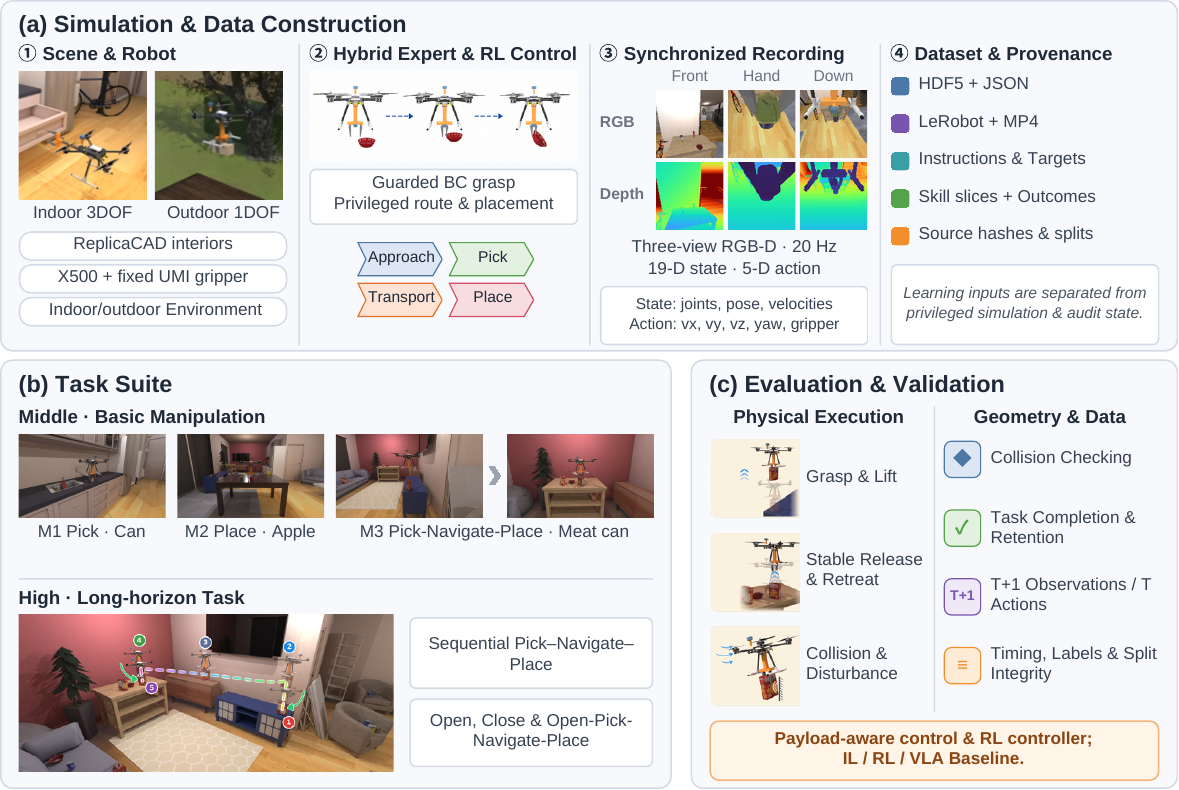}
    \caption{
        Overview of \methodname{}.
        (a) \textbf{Simulation and data construction:}
        indoor and outdoor aerial manipulation scenes are instantiated with
        X500-based platforms, while hybrid expert and RL policies generate
        manipulation behaviors that are synchronously recorded as multi-view
        RGB-D observations, robot states, actions, task annotations, and
        provenance metadata.
        (b) \textbf{Task suite:}
        the benchmark covers basic manipulation tasks, including
        \textsc{Pick}, \textsc{Place}, and
        \textsc{Pick--Navigate--Place}, as well as long-horizon tasks such as
        sequential pick--navigate--place and
        \textsc{Open--Pick--Navigate--Place}.
        (c) \textbf{Evaluation and validation:}
        trajectories are evaluated through physical execution criteria,
        collision and disturbance checks, task-completion validation, and
        temporal/data-integrity checks, supporting IL, RL, and VLA baselines.
    }
    \label{fig:benchmark_overview}
\end{figure}

Figure~\ref{fig:benchmark_overview} summarizes the overall design of
\methodname{}.
The benchmark integrates simulation, automated data generation, task
construction, and systematic evaluation within a unified pipeline.
Manipulation trajectories are generated using privileged hybrid experts
and reinforcement learning controllers, and are recorded with synchronized
multi-view RGB-D observations, robot states, actions, task annotations, and
provenance metadata.
The resulting task suite spans both short-horizon manipulation and
long-horizon navigation--manipulation compositions.
Finally, each trajectory is validated through physical execution,
collision checking, task-completion criteria, and temporal consistency
checks before being used for IL, RL, or VLA training and evaluation.

\subsubsection{Robot platforms}
\label{app:platforms}

\begin{figure}[t]
    \centering
    \includegraphics[width=\linewidth]{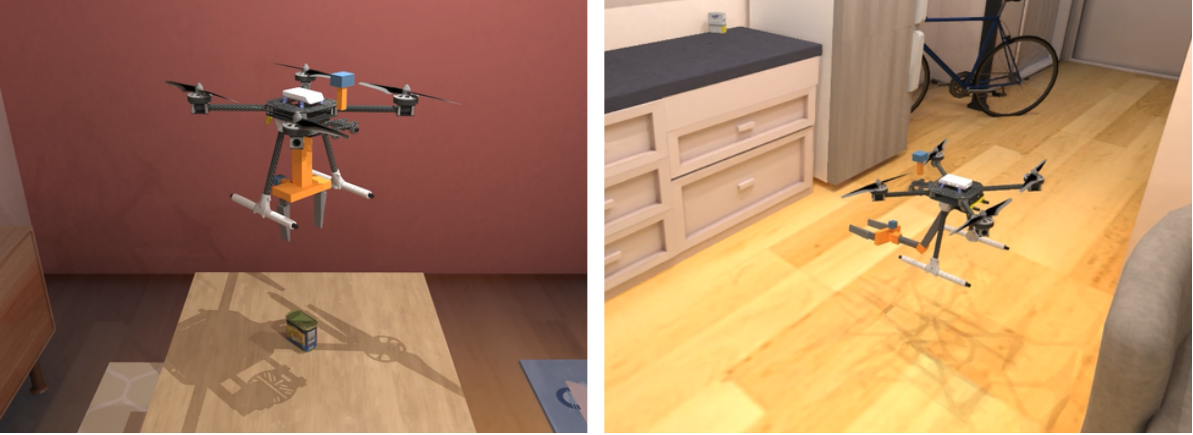}
    \caption{
        Aerial manipulation platforms in \methodname{}.
        Left: an X500-based platform with a downward-facing
        gripper mounted beneath the airframe.
        Right: an X500-based platform with a forward-reaching,
        arm-mounted gripper.
        Both platforms use the same flight-control interface
        and carry three onboard RGB-D cameras.
    }
    \label{fig:platforms}
\end{figure}

\methodname{} supports two aerial manipulation platforms with
different gripper configurations (Fig.~\ref{fig:platforms}).
The first carries a downward-facing gripper beneath the airframe,
while the second uses a forward-reaching, arm-mounted gripper.
Both platforms share the PX4 X500 airframe, flight-control
interface, and simulation setup.

Both platforms have a rotor arm length of $0.174$\,m and a
rotor-disk radius of $0.147$\,m, and are simulated in
ManiSkill/SAPIEN with physics at $240$\,Hz and control at $20$\,Hz.
The shared flight interface accepts world-frame linear velocity
and yaw-rate commands, bounded by $0.6$\,m/s and $1$\,rad/s,
respectively.
A PI velocity controller computes a desired acceleration,
from which the desired thrust and attitude are determined.
A geometric PD attitude controller then computes the desired
body torque.
The thrust and torque commands are allocated to individual
rotors subject to per-rotor saturation, and the resulting net
force and torque are applied to the airframe.
Each platform carries three onboard cameras providing
$128\times128$ RGB-D observations.
Table~\ref{tab:platforms} summarizes the platform parameters.

\paragraph{Gripper platform.} A one-degree-of-freedom parallel gripper with the long UMI
fingertips~\citep{UMI} is mounted rigidly below the airframe; the grasp point is 0.28\,m below
the airframe centre. The X500 mass and inertia are kept from the source model; the adapter (100\,g),
palm (80\,g), cameras and mounts (80\,g) and fingers (30\,g each) are nominal values.

\paragraph{Arm platform.} Handles on vertical surfaces lie outside the reach of the gripper platform,
because its rotor disks extend 0.32\,m beyond the grasp point in every horizontal direction. The arm
platform removes the gripper mount and adds a three-degree-of-freedom arm under the frame: shoulder
pitch and elbow pitch about the body $y$ axis, then wrist roll about the forearm. The shoulder axis is
2\,cm forward of and 4.5\,cm below the airframe centre. In the working pose (upper arm pitched up
0.12\,rad, forearm level) the grasp point is 0.47\,m ahead of and 2\,cm below the centre. At zero roll
the fingers close horizontally, which fits the 50\,mm-wide vertical bar on the fridge door; at 90° they
close vertically, which fits the 24\,mm-tall drawer handle.

Three parameters differ from the gripper platform, and each is set by a measured requirement. Starting
the fridge door moving takes about 7\,N at the handle, and the drawer about 2.5\,N. With the gripper
platform's finger drive (150\,N/m) a pinch on these handles reaches only 2--4\,N, so the finger drive is
raised to 1500\,N/m with a 15\,N limit per finger. With the stock rotors (8.55\,N) the vehicle saturates
at the 20--25° tilt it needs to pull the door, so the motors are upgraded to 11.5\,N per rotor (a nominal
150\,g added at the motor mounts). The flight loop adds a feedforward torque that cancels the arm's weight
moment about the airframe centre of mass (about 0.7\,N\,m with the arm extended). It also uses stiffer
attitude gains, a higher horizontal acceleration limit, and a 0.12\,N\,m limit on yaw torque: a grasped
door sets the heading, and an unlimited yaw error would saturate all four rotors.

\paragraph{Contact mode.} A two-finger pinch on a bar acts as a hinge about the finger-closing axis,
0.47\,m ahead of the centre of mass. With a stiff shoulder, the vehicle can only tilt to generate the
pulling force by swinging about that hinge. While a handle is pinched, the shoulder servo is therefore
set to zero torque and the arm feedforward is disabled; the arm then acts as a strut that transmits force
at the shoulder. On release, the shoulder is stiffened again and the velocity integrator is reset.

\paragraph{Modelling assumptions.} Arm link masses (0.17\,kg in total, plus an 82\,g shoulder servo) and
joint torque limits are nominal hobby-servo figures, not measurements of a built arm. Links and contacts
are rigid, with no flexible fingers, motor lag, propeller wash or ground effect.

\begin{table}[t]
\centering
\small
\caption{Parameters of the two platforms. Both use the same airframe, flight interface and camera
resolution; the right column lists only what the arm platform adds or changes.}
\label{tab:platforms}
\begin{tabular}{lll}
\toprule
 & Gripper platform & Arm platform \\
\midrule
Total mass $m_0$ & 2.38\,kg & 2.58\,kg \\
Max thrust per rotor & 8.55\,N & 11.5\,N \\
Thrust-to-weight ratio & 1.46 & 1.82 \\
Manipulator & fixed 1-DOF gripper & 3-DOF arm + 1-DOF gripper \\
Shoulder pitch & -- & $[-0.6, 1.9]$\,rad, 3.0\,N\,m \\
Elbow pitch & -- & $\pm 2.6$\,rad, 0.9\,N\,m \\
Wrist roll & -- & $\pm 1.75$\,rad, 0.5\,N\,m \\
Link lengths & -- & 0.22\,m / 0.12\,m / 0.12\,m to grasp point \\
Grasp point (body frame) & 0.28\,m below centre & 0.47\,m ahead, 0.02\,m below \\
Fingers & UMI long, 120\,mm, 0--110\,mm & same \\
Finger drive & 150\,N/m, 5\,N per finger & 1500\,N/m, 15\,N per finger \\
Hand camera & below frame, looks down at jaw & on wrist, looks along fingers \\
Down camera & under frame, looks down-forward & under right front, side view \\
Action dimension & 4 flight + 1 gripper & 4 flight + 3 arm + 1 gripper \\
Attitude gains $k_p$ / $k_d$ & 2.5 / 0.55 & 6 / 1.0 \\
Horizontal accel.\ limit & 3\,m/s$^2$ & 4.5\,m/s$^2$ \\
Velocity-integral limit & 2\,m/s$^2$ & 4\,m/s$^2$ \\
Extra flight terms & -- & arm feedforward, yaw torque $\le$ 0.12\,N\,m \\
Target subtasks & pick, place & open, close (fridge, counter drawer) \\
\bottomrule
\end{tabular}
\end{table}

\subsection{Trajectory Annotation and Outcome Categorization}
\label{sec:appendix:trajectory_modes}

We characterize trajectory outcomes using simulator-derived progress
indicators, failure flags, and task-specific completion checks.
Tables~\ref{tab:pick_modes}--\ref{tab:close_modes} summarize the
diagnostic modes used for aerial \textsc{Pick}, \textsc{Place},
\textsc{Open}, and \textsc{Close}. Within each table, the first
satisfied row determines the primary reported mode, while all underlying
diagnostic signals are retained. Initialization failures and infeasible
starts are recorded separately from executed skill trajectories.

\paragraph{Pick/Place Outcome Semantics and Completion Checks.}

For \textsc{Pick} and \textsc{Place}, $S$, $F$, and $T$ denote
successful completion, physical failure, and time-limit truncation,
respectively, while $c$ denotes the recorded primary failure cause.

For any recorded binary event predicate $b_t$, we define its first
occurrence as
\[
    t_b = \min\{t:b_t=1\},
\]
with $t_b=\infty$ if the event is never observed.

For \textsc{Pick}, $t_g$ and $t_l$ denote the first grasp and lift
events. For \textsc{Place}, $t_o$, $t_d$, $t_r$, and $t_c$ denote
the first overhead-alignment, support-contact, release, and
gripper-clearance events, respectively.

These event times are treated as independent observations rather than
an assumed monotonic stage sequence. Time-limit truncations indicate
incomplete execution and are not relabeled as physical failures.

A \textsc{Pick} succeeds when the object is grasped and maintained above
its instance-specific lift threshold for the required consecutive
control steps.
A \textsc{Place} succeeds when the placement predicate is satisfied,
the gripper is open, release has been registered, the gripper is more
than $0.27\,\mathrm{m}$ from the object, and platform speed remains
below $0.025\,\mathrm{m/s}$ for 20 consecutive control steps without
a latched protocol failure. The placement predicate jointly evaluates
the goal region, object stability, and support-contact or resting-height
conditions.

\begin{table*}[t]
\centering
\small
\renewcommand{\arraystretch}{1.15}
\caption{
Outcome categories for \textsc{Pick}.
$c$ denotes the recorded primary failure cause.
}
\label{tab:pick_modes}
\begin{tabularx}{\textwidth}{@{}p{0.29\textwidth}X@{}}
\toprule
Mode & Definition \\
\midrule
Successful grasp handoff
& $S$: the Pick completion condition is satisfied. \\

Collision failure
& $F$ and $c=\texttt{collision}$: the UAV makes unintended
contact with non-target scene geometry during execution. \\

Object-drop failure
& $F$ and $c=\texttt{fail\_object\_drop}$: the object violates the
minimum-height condition. \\

Flight-envelope failure
& $F$ and
$c\in\{\texttt{fail\_tilt},\texttt{fail\_low},\texttt{fail\_high}\}$:
the UAV exceeds the allowed attitude or altitude envelope, corresponding
to excessive tilt, altitude below the lower bound, or altitude above
the upper bound, respectively. \\

Other failure
& $F$ with any remaining cause, including workspace violations or
unclassified failures. \\

No-grasp truncation
& $T$ and $t_g=\infty$: no grasp is observed before timeout. \\

Grasp-without-lift truncation
& $T$, $t_g<\infty$, and $t_l=\infty$: a grasp is observed, but the
lift threshold is never reached. \\

Incomplete-handoff truncation
& $T$, $t_g<\infty$, and $t_l<\infty$: grasp and lift are observed,
but the final handoff condition is not completed. \\
\bottomrule
\end{tabularx}
\end{table*}

\begin{table*}[t]
\centering
\small
\renewcommand{\arraystretch}{1.15}
\caption{
Outcome categories for \textsc{Place}, defined from recorded progress
events and task-specific failure signals.
}
\label{tab:place_modes}
\begin{tabularx}{\textwidth}{@{}p{0.29\textwidth}X@{}}
\toprule
Mode & Definition \\
\midrule
Successful placement
& $S$: placement, release, clearance, and sustained platform stability
satisfy the protocol. \\

Collision-budget failure
& $F$ with primary cause \texttt{fail\_collision\_force}: the accumulated
collision-force score exceeds its configured budget. \\

Flight-envelope failure
& $F$ with primary cause
\texttt{fail\_tilt}, \texttt{fail\_low}, or \texttt{fail\_high}. \\

Workspace failure
& $F$ with primary cause \texttt{fail\_workspace}: the UAV position
violates the predefined workspace boundary during execution. \\

Object-drop failure
& $F$ with primary cause \texttt{fail\_object\_drop}: object height falls
below the floor or support-relative threshold. \\

Unsupported-release failure
& $F$ with primary cause \texttt{fail\_unsupported\_release}: grasp loss
persists for the configured duration without sufficient support evidence
and precedes an accepted release. \\

Other failure
& $F$ with any remaining cause, including non-finite policy features. \\

No-alignment truncation
& $T$ and $t_o=\infty$: the object never enters the horizontal alignment
tolerance. \\

No-support truncation
& $T$, $t_o<\infty$, and $t_d=\infty$: alignment is observed, but support
contact inside the goal region is not. \\

No-release truncation
& $T$, $t_o,t_d<\infty$, and $t_r=\infty$: the release event is not
observed. \\

Incomplete-completion truncation
& Remaining $T$ cases in which the observed events do not satisfy the
complete placement, clearance, and stability conditions. \\
\bottomrule
\end{tabularx}
\end{table*}

\paragraph{Open/Close Outcome Semantics and Completion Checks.}

For \textsc{Open} and \textsc{Close}, let $q$ denote the current
articulation position, with limits $q_{\min}$ and $q_{\max}$. We define
the normalized articulation progress as
\[
    \bar q =
    \frac{q-q_{\min}}{q_{\max}-q_{\min}}.
\]

Opening requires $\bar q\geq\alpha_a$, where $\alpha_a$ denotes the
articulation-specific opening threshold
($0.75$ for the refrigerator and $0.90$ for the kitchen-counter
drawer), while closing requires $\bar q\leq0.01$.

Let $p$ denote the terminal execution phase. We use the binary checks
$J$, $R$, $A$, $V$, $N$, and $C$ to denote articulation-goal
attainment, handle release, arm retraction, platform stability,
absence of airframe contact, and geometric clearance, respectively.
The overall articulation success indicator is

\[
    S_{\mathrm{art}}
    =
    J \land R \land A \land V \land N \land C
    \land [p=\texttt{done}].
\]

Here, $S_{\mathrm{art}}$ denotes successful completion of the
articulated manipulation skill.

Arm retraction requires a maximum joint deviation below
$0.05\,\mathrm{rad}$ from the prescribed rest pose.
Platform stability requires linear speed below
$0.05\,\mathrm{m/s}$ and angular speed below
$0.2\,\mathrm{rad/s}$.
The release condition is satisfied when the two fingers no longer both
exceed the $0.1\,\mathrm{N}$ contact-force threshold.
The clearance check requires a minimum sampled distance of
$0.03\,\mathrm{m}$ between the robot geometry, excluding the fingers,
and the monitored furniture geometry throughout the rollout.

\begin{table*}[t]
\centering
\small
\renewcommand{\arraystretch}{1.15}
\caption{
Outcome categories for \textsc{Open}, based on articulation progress,
completion checks, and recorded execution phases.
}
\label{tab:open_modes}
\begin{tabularx}{\textwidth}{@{}p{0.29\textwidth}X@{}}
\toprule
Mode & Definition \\
\midrule
Successful opening
& $S_{\mathrm{art}}$: the opening threshold and all completion checks
are satisfied. \\

Airframe-contact failure
& $\neg N$: at least one airframe contact is recorded. \\

Clearance-check failure
& $\neg C$: the minimum sampled geometric clearance is below
$0.03\,\mathrm{m}$. \\

Handle-approach failure
& $p=\texttt{fail\_insert}$: the approach stage fails to satisfy its
target condition within the phase limit. \\

Handle-pinch failure
& $p=\texttt{fail\_pinch}$: the required finger-contact condition is
not satisfied. \\

Handle-slip failure
& $p=\texttt{fail\_slip}$: handle contact is lost before the opening
goal is reached. \\

Opening-drive failure
& $p=\texttt{fail\_drive}$: the drive phase exceeds its time limit
before satisfying its transition condition. \\

Opening-goal unmet
& $\neg J$: the final articulation position does not satisfy
$\bar q\geq\alpha_a$. \\

Handle-release incomplete
& $\neg R$: the terminal release condition is not satisfied. \\

Arm-retraction incomplete
& $\neg A$: the arm does not return to the prescribed rest pose. \\

Platform-settling incomplete
& $\neg V$: terminal linear or angular velocity exceeds the stability
threshold. \\

Execution incomplete
& Remaining unsuccessful cases with $p\neq\texttt{done}$. \\
\bottomrule
\end{tabularx}
\end{table*}

\begin{table*}[t]
\centering
\small
\renewcommand{\arraystretch}{1.15}
\caption{
Outcome categories for \textsc{Close}, using the closing threshold and
the same completion and safety checks as \textsc{Open}.
}
\label{tab:close_modes}
\begin{tabularx}{\textwidth}{@{}p{0.29\textwidth}X@{}}
\toprule
Mode & Definition \\
\midrule
Successful closing
& $S_{\mathrm{art}}$: the closing threshold and all completion checks
are satisfied. \\

Airframe-contact failure
& $\neg N$: at least one airframe contact is recorded. \\

Clearance-check failure
& $\neg C$: the minimum sampled geometric clearance is below
$0.03\,\mathrm{m}$. \\

Handle-approach failure
& $p=\texttt{fail\_insert}$: the approach stage fails to satisfy its
target condition within the phase limit. \\

Handle-pinch failure
& $p=\texttt{fail\_pinch}$: the required finger-contact condition is
not satisfied. \\

Handle-slip failure
& $p=\texttt{fail\_slip}$: handle contact is lost before the closing
goal is reached. \\

Closing-drive failure
& $p=\texttt{fail\_drive}$: the drive phase exceeds its time limit
before satisfying its transition condition. \\

Closing-goal unmet
& $\neg J$: the final articulation position does not satisfy
$\bar q\leq0.01$. \\

Handle-release incomplete
& $\neg R$: the terminal release condition is not satisfied. \\

Arm-retraction incomplete
& $\neg A$: the arm does not return to the prescribed rest pose. \\

Platform-settling incomplete
& $\neg V$: terminal linear or angular velocity exceeds the stability
threshold. \\

Execution incomplete
& Remaining unsuccessful cases with $p\neq\texttt{done}$. \\
\bottomrule
\end{tabularx}
\end{table*}

\paragraph{Demonstration Selection and Skill Attribution.}

The recorded modes support task-specific demonstration selection and
structured analysis of unsuccessful executions. Failed and truncated
trajectories can be retained together with successful demonstrations,
depending on the export configuration.

For segmented trajectories, skill-level labels are stored separately
from the diagnostics of the original parent rollout. A completed skill
therefore remains valid if a later stage fails, while concurrent
physical failures are preserved as failures. Place segments are
additionally required to begin from an eligible held state outside the
destination support region.


%

\end{document}